\pdfoutput=1
\documentclass[10pt,twocolumn,letterpaper]{article}
\usepackage[letterpaper,margin=0.75in,columnsep=0.25in]{geometry}
\usepackage[T1]{fontenc}
\usepackage{amsmath}
\usepackage{newtxtext,newtxmath}
\usepackage{courier}
\usepackage[hyphens]{url}
\usepackage{graphicx}
\usepackage{natbib}
\usepackage{caption}
\usepackage{booktabs}
\usepackage{multirow}
\usepackage[table]{xcolor}
\usepackage{makecell}
\usepackage{colortbl}
\usepackage{newfloat}
\usepackage{listings}
\usepackage[colorlinks,allcolors=blue!60!black,breaklinks]{hyperref}
\usepackage{algorithm}
\usepackage{algpseudocode}
\definecolor{wafirrow}{RGB}{255,255,180}
\definecolor{bestred}{RGB}{220,0,0}
\definecolor{secondblue}{RGB}{0,70,200}

\newcommand{\suppfig}[1]{Supplementary Figure~S#1}
\newcommand{\supptab}[1]{Supplementary Table~S#1}
\newcommand{\suppalg}[1]{Supplementary Algorithm~S#1}
\newcommand{\suppsec}[1]{Supplementary Section~S#1}
\title{WaFIR: Wavelet-Fourier Mixture-of-Experts with Subband-Adaptive Routing for Blind All-in-One Image Restoration}
\author{%
Trong Nhan Nguyen$^{1}$ \quad Vu Hoang Minh Doan$^{1}$ \quad Thanh Dat Le$^{1}$ \quad Truong Tien Vo$^{1}$ \quad Thi Thu Ha Vu$^{1}$\\
Tan Hung Vo$^{1}$ \quad Jaeyeop Choi$^{1}$ \quad Junghwan Oh$^{1,*}$ \quad Truong-Son Hy$^{2,*}$\\[6pt]
{\normalsize $^{1}$Pukyong National University, Busan, Republic of Korea}\\
{\normalsize $^{2}$University of Alabama at Birmingham, Birmingham, AL, USA}\\[4pt]
{\normalsize $^{*}$Corresponding authors: Junghwan Oh (\href{mailto:jungoh@pknu.ac.kr}{\texttt{jungoh@pknu.ac.kr}}), Truong-Son Hy (\href{mailto:thy@uab.edu}{\texttt{thy@uab.edu}})}}
\date{}
\begin{document}
\maketitle

\begin{abstract}
Blind all-in-one image restoration must adapt a shared model to heterogeneous and sometimes mixed degradations without a task label at test time. We introduce WaFIR, a decoder module that decomposes features into Haar subbands, summarizes each subband with a compact Fourier descriptor, and routes it to prompt-conditioned spectral modulation experts. Unlike image-level frequency conditioning, WaFIR makes routing decisions at the subband level, enabling frequency-specific processing while retaining the task context learned by the backbone. Under a unified training and evaluation protocol for denoising, deraining, and dehazing, WaFIR improves average PSNR over the three tasks by 0.48 dB over the PromptIR baseline and by 0.38 dB over the recent Up-Restorer. Matched-capacity and routing controls show that the gain is not explained by additional parameters alone. On unseen noise intensities, descriptor-based routing is more robust than a discrete label-conditioned alternative and remains on par under compound degradations, while routing analyses show stable subband-dependent expert use across degradation types. These results support subband frequency structure as an interpretable complement to prompt-based conditioning for blind image restoration. Code is available at \url{https://anonymous.4open.science/r/WaFIR/README.md}.
\end{abstract}

\section{Introduction}
\label{sec:intro}

\begin{figure*}[t]
  \centering
  \includegraphics[width=0.9\linewidth]{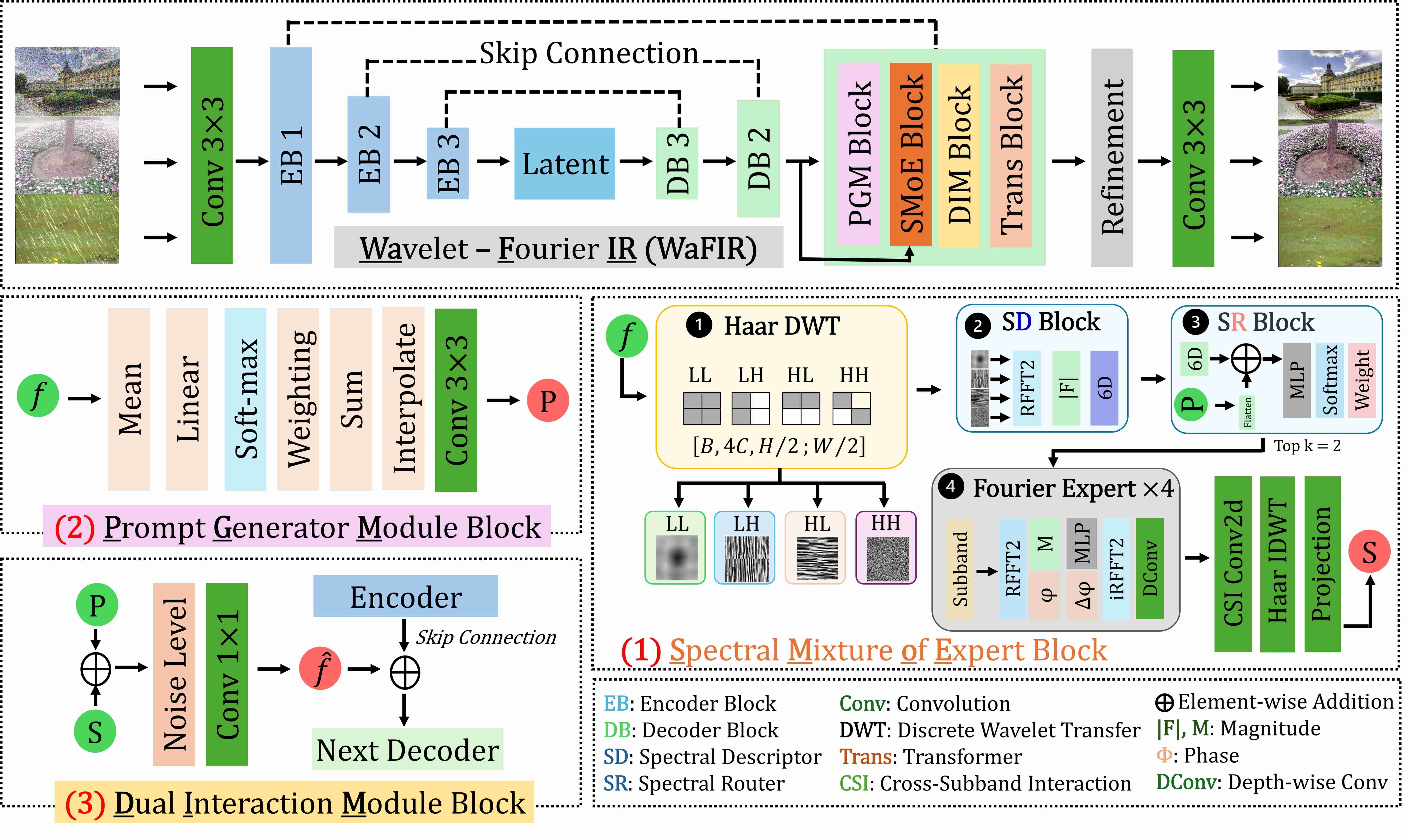}
  \caption{%
  \textbf{Overall architecture of WaFIR.} A PromptIR-style U-Net is augmented at each
  decoder level by three modules. \textbf{(1) SPMoE} decomposes the feature
  via Haar DWT into directional subbands (LL/LH/HL/HH) for multi-resolution
  \emph{spatial--scale} localization, describes each with a $6$-D Fourier
  descriptor (SD), routes it to top-$k$ Fourier Experts (SR) that modulate
  magnitude and phase in the \emph{global} frequency domain, and reconstructs
  via cross-subband interaction (CSI) and Haar IDWT. Wavelet and FFT are
  complementary: the wavelet gives spatial locality a global FFT lacks; the
  FFT gives frequency (magnitude+phase) selectivity a wavelet subband lacks. 
  \textbf{(2) PGM} generates a degradation-adaptive prompt $P$; \textbf{(3) DIM}
  fuses $P$ with the spectral output $S$ and the skip connection for the next
  decoder.%
  }
  \label{fig:overview}
\end{figure*}

\begin{figure}[!t]
  \centering
  \includegraphics[width=0.92\linewidth]{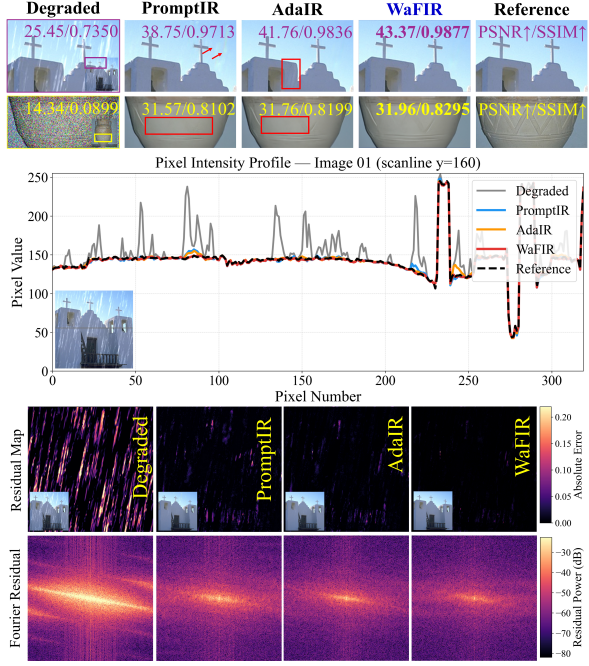}
  \caption{%
    \textbf{Qualitative comparison and frequency-domain error analysis.}
  \emph{Rows~1--2:} derain and denoise ($\sigma{=}50$) examples; columns are
  degraded input, PromptIR, AdaIR, \textbf{WaFIR (ours)}, and reference, with
  PSNR\,/\,SSIM below each patch. On derain (row~1) WaFIR improves
  $+1.61$\,dB over AdaIR (ICLR'25) and $+4.62$\,dB over PromptIR (NeurIPS'23);
  on denoise (row~2), beyond the higher scores, the zoom crops show WaFIR
  better preserves fine texture that the baselines over-smooth.
  \emph{Row~3:} 1D intensity profile ($y{=}160$); WaFIR (red) tracks the
  reference (black dashed) most closely at edges. \emph{Rows~4--5:} residual
  maps $|\hat{x}-x_{\text{gt}}|$ and their Fourier spectra. Rain leaves a
  directional high-frequency band in the residual; WaFIR suppresses it most,
  with the steepest spectral energy decay, confirming removal of
  degradation-specific frequency structure.
  }
  \label{fig:qual_rain}
\end{figure}

Real-world images suffer from diverse degradations such as sensor noise, haze and rain streak. Classical restoration handles these with separate task-specific models~\cite{dncnn,mprnet,retinexnet}, an approach that scales poorly as the number of tasks grows. All-in-one restoration instead recovers multiple degradations with a single model~\cite{airnet,transweather,promptir,adair,moceir}, where the central question is how to adapt computation to each input.
Existing methods answer this with different conditioning signals, including contrastive degradation embeddings~\cite{airnet}, learnable prompts~\cite{promptir}, physics-inspired latent variables~\cite{uprestorer}, and uncertainty-based task weighting~\cite{tur}. Despite architectural differences, all of them condition adaptation on a representation that summarizes the feature map without resolving its spectral organization. Such signals identify the overall restoration context but carry little information about how degradation energy is distributed across frequency subbands, so feature regions with distinct spectral characteristics may receive similar routing decisions.
This limitation matters most when models must generalize beyond their training conditions. Degradations that produce similar global representations often differ sharply in spectral distribution. Haze alters low-frequency components, rain streaks concentrate directional energy in middle-to-high-frequency subbands, and additive noise, though approximately uniform across the spectrum, dominates high frequencies because natural-image spectra decay with frequency. These characteristics are more naturally distinguished in frequency subspaces than by global summaries, and because they are computable directly from intermediate features (\suppfig{1}), they remain available even when degradation types or intensities are unseen at test time.
We therefore ask whether \emph{subband-level spectral structure} provides a more informative routing signal than spectrally unresolved representations, with prompt features supplying complementary task context. Building on this, we propose \textbf{WaFIR (Wavelet--Fourier Image Restoration)}, which augments PromptIR~\cite{promptir} with a \textbf{Spectral Mixture-of-Experts (SPMoE)} module at each decoder stage. SPMoE decomposes each decoder feature into four Haar subbands (LL, LH, HL, HH), extracts a compact six-dimensional Fourier descriptor per subband, and routes each subband through a sparse top-$k$ gate to specialized Fourier experts that perform prompt-conditioned magnitude modulation. Routing is conditioned jointly on the descriptor and the pooled prompt embedding. Wavelets localize \emph{where} and \emph{at what scale} spectral information sits, whereas Fourier modulation adapts \emph{which frequencies} are corrected.
Analyses support this principle. Routing weights organize primarily by wavelet subband while carrying only chance-level linearly decodable degradation information, indicating that experts self-organize by frequency subspace rather than degradation identity. Under unseen noise intensities, descriptor and prompt routing generalizes more robustly than discrete label-conditioned routing, and it remains on par or better under compound degradations.
Our contributions are:
\begin{enumerate}
\item \textbf{Subband-level spectral routing.} We introduce per-subband Fourier descriptors as an input-derived routing signal for unified restoration, jointly conditioning sparse expert routing with pooled prompt features.
\item \textbf{Wavelet-Fourier expert design.} We propose the SPMoE module, pairing Haar decomposition with prompt-conditioned Fourier spectral-modulation experts, stabilized by a zero-initialized residual gate and a per-subband load-balancing loss.
\item \textbf{Comprehensive evaluation and interpretable routing.} We show that WaFIR improves the PromptIR backbone on denoising and deraining, is competitive with the state of the art overall, and generalizes without fine-tuning to unseen datasets and an industrial domain. We further show that its experts self-organize by frequency subband, and that descriptor-based routing is more robust than a label-conditioned alternative under unseen noise intensities.
\end{enumerate}

\section{Related Work}
\label{sec:related}
\noindent\textbf{Blind All-in-One Image Restoration:}
Unified restoration methods recover multiple degradation types with a single network by learning adaptive conditioning signals from the input. PromptIR~\cite{promptir} extracts learnable prompts from decoder features, AdaIR~\cite{adair} mines and modulates low- and high-frequency feature components, and more recent methods such as OneRestore~\cite{guo2024onerestore} and FAPE-IR~\cite{fapeir} strengthen feature adaptation for harder degradation scenarios. Despite these differences, adaptation is applied at a coarse granularity---a global conditioning signal or a two-way low/high-frequency split---so fine-grained frequency regions within the same feature map are not resolved into separately specialized computations. WaFIR instead conditions after resolving the feature into wavelet subbands, letting different spectral regions receive different expert assignments while remaining jointly conditioned by prompt context.

\noindent\textbf{Mixture-of-Experts for Adaptive Restoration:}
MoE enables adaptive computation by activating a sparse subset of experts per input~\cite{shazeer2017outrageously}. Restoration models such as MoCE-IR~\cite{moceir} and MOERL~\cite{wang2025moerl} route experts by degradation-aware embeddings, restoration complexity, or global feature representations, organizing specialization at the feature level. WaFIR instead asks whether the frequency organization of the feature provides a better basis for specialization: rather than assigning one expert to an entire feature, each wavelet subband is routed independently by its own spectral descriptor, so experts specialize on complementary frequency structures within the same image.

\noindent\textbf{Frequency-aware Image Restoration:}
Frequency representations are widely used in restoration. MWCNN~\cite{mwcnn} applies wavelet decomposition for multi-scale representation, and FFC~\cite{ffc} introduces Fourier convolutions to enlarge the receptive field. Among unified methods, AdaIR~\cite{adair} is closest to our work: it likewise operates in the frequency domain, mining low- and high-frequency components from the features (guided by the degraded image's spectra) and modulating them through a bidirectional operator. We take a different approach to processing the frequency domain: WaFIR extracts per-subband Fourier descriptors over four Haar subbands and uses them directly for sparse top-$k$ routing, so different spectral regions activate different experts. This is a frequency-resolved routing mechanism rather than a global frequency-modulation stage.

Prompt conditioning, frequency representations, and sparse expert routing each improve unified restoration independently. WaFIR connects them through a different question: \emph{can frequency subbands themselves serve as an interpretable basis for expert specialization?} Rather than combining existing modules, WaFIR shows that routing self-organizes by wavelet subband rather than degradation category, giving a frequency-grounded routing principle that generalizes more robustly to unseen degradations.

\section{Method}
\label{sec:method}

\subsection{Overview}

WaFIR builds on PromptIR~\cite{promptir}, a hierarchical encoder--decoder
whose decoder levels are conditioned on learnable degradation prompts.
We insert a \textbf{SpectralMoEBlock} $\mathcal{S}_\ell$ immediately after
the prompt generation step at each of the three decoder levels
$\ell \in \{3,2,1\}$:
\begin{equation}
  \mathbf{f}'_\ell = \mathcal{S}_\ell(\mathbf{f}_\ell,\, \mathbf{p}_\ell),
  \label{eq:overview}
\end{equation}
where $\mathbf{f}_\ell \in \mathbb{R}^{B\times C_\ell\times H_\ell\times W_\ell}$
is the decoder feature, $\mathbf{p}_\ell$ is the prompt, and
$C_\ell \in \{384,192,96\}$.
The enriched feature $\mathbf{f}'_\ell$ is then concatenated with $\mathbf{p}_\ell$
and fed to the standard PromptIR transformer block.
All other backbone components are unchanged.
Setting $\mathcal{S}_\ell$ to identity recovers vanilla PromptIR exactly,
making SPMoE fully plug-and-play. Figure~\ref{fig:overview} illustrates the overall pipeline, while the complete forward pseudocode of the proposed SpectralMoEBlock is given in \suppalg{1}--\suppalg{2}.

\subsection{Haar Wavelet Decomposition}

Given a feature map $\mathbf{x} \in \mathbb{R}^{B\times C\times H\times W}$,
a single-level 2-D Haar DWT~\cite{mallat2009} decomposes it into four subbands by applying
separable averaging ($\mathbf{L}$) and differencing ($\mathbf{H}$) filters
along both spatial axes:
\begin{equation}
  \mathrm{DWT}(\mathbf{x}) =
  [\mathbf{LL};\mathbf{LH};\mathbf{HL};\mathbf{HH}]
  \in \mathbb{R}^{B\times 4C\times H/2\times W/2}.
  \label{eq:dwt}
\end{equation}
The normalization of $\tfrac{1}{\sqrt{2}}$ at each stage makes the transform
unitary ($\|\mathrm{DWT}(\mathbf{x})\|^2 = \|\mathbf{x}\|^2$), so the
inverse IDWT reconstructs $\mathbf{x}$ exactly without loss of information.
The Haar DWT does not require trainable parameters and runs in $O(N)$ time.

According to \suppfig{1}, each
 subband carries a physically interpretable frequency signature
that naturally aligns with a specific degradation type, as summarized in
\supptab{1}.

\subsection{Spectral Descriptor}

For each subband $\mathbf{s} \in \mathbb{R}^{B\times C\times H_s\times W_s}$,
a parameter-free \textbf{SpectralDescriptor} computes a 6-dimensional
summary of its 2-D magnitude spectrum.
Let $\mathbf{M}(u,v) = |\mathrm{RFFT2}(\mathbf{s})| \in
\mathbb{R}^{B\times C\times H_s\times (W_s/2{+}1)}$ be the
orthonormal real-FFT magnitude, and define the low-frequency boundary
$l_H = \lfloor H_s \cdot 0.25\rfloor$, $l_W = \lfloor W_s/2 \cdot 0.25\rfloor$.
The descriptor $\mathbf{d} \in \mathbb{R}^{B\times 6}$ is:
\begin{align}
  d_0 &= \textstyle\overline{M}_{u,v,c}   & &\text{(global energy)} \label{eq:d0}\\
  d_1 &= \mathrm{Var}_{u,v,c}[M]          & &\text{(spectral spread)} \label{eq:d1}\\
  d_2 &= \overline{M}_{u<l_H,\,v<l_W}     & &\text{(DC / low-freq)} \label{eq:d2}\\
  d_3 &= \overline{M}_{u\geq l_H,\,v\geq l_W} & &\text{(high-freq)} \label{eq:d3}\\
  d_4 &= \overline{M}_{l_H\,\leq u<\,H_s/2} & &\text{(horiz.\ band)} \label{eq:d4}\\
  d_5 &= \overline{M}_{l_W\,\leq v\leq\,W_s/2} & &\text{(vert.\ band)} \label{eq:d5}
\end{align}
Each dimension captures a degradation-discriminative spectral property:
$d_0$ is elevated by haze; $d_1$ is high for noise (flat spectrum);
$d_2$ responds to haze; $d_3$ to noise and sharp edges;
$d_4$--$d_5$ to rain streaks and directional blur.
The descriptor is computed independently for all four subbands, resulting
in $\mathbf{d}^{\text{LL}}, \mathbf{d}^{\text{LH}}, \mathbf{d}^{\text{HL}},
\mathbf{d}^{\text{HH}} \in \mathbb{R}^{B\times 6}$.

\subsection{Spectral Router and Sparse Top-$k$ Gating}

The router for each subband fuses the spectral descriptor with a pooled
prompt vector:
\begin{equation}
  \mathbf{r} = [\mathbf{d};\,\bar{\mathbf{p}}],
  \quad \bar{\mathbf{p}} = \mathrm{AvgPool}(\mathbf{p}) \in \mathbb{R}^{B\times C_p}.
  \label{eq:router_input}
\end{equation}
A two-layer MLP maps $\mathbf{r}$ to $E{=}4$ expert logits:
\begin{equation}
  \mathbf{g} = W_2\,\mathrm{GELU}(W_1 \mathbf{r}) \in \mathbb{R}^{B\times E},
  \label{eq:router_mlp}
\end{equation}
followed by sparse top-$k$ selection ($k{=}2$) and softmax normalization:
\begin{equation}
  w_i =
  \begin{cases}
    \mathrm{softmax}(\mathbf{g})_i & \text{if } g_i \geq g_{(k)}, \\
    0 & \text{otherwise},
  \end{cases}
  \label{eq:topk}
\end{equation}
where $g_{(k)}$ is the $k$-th largest logit.
Only the $k$ selected experts contribute to the output, keeping the
per-sample, per-subband expert computation fixed as $E$ grows.
The descriptor $\mathbf{d}$ provides subband-intrinsic spectral statistics;
the prompt $\bar{\mathbf{p}}$ encodes task context.
No degradation type label is required at inference.

\noindent\textbf{Roles of the two conditioning signals:}
The gate is \emph{jointly} conditioned on $\mathbf{d}$ and
$\bar{\mathbf{p}}$; the architecture itself therefore does not enforce a
separation of roles, and we characterize the roles empirically instead.
Ablations (Table~\ref{tab:ablation}; expanded discussion in \suppsec{3}) show that a descriptor-only gate
(A14, $-0.15$\,dB) retains more of the full model's performance than a
prompt-only gate (A12, $-0.26$\,dB), indicating that the intrinsic physical characteristics reflected in the subband spectrum already contain useful routing information, while the prompt provides complementary task-context information.
Consistently, the quantitative routing analysis in \suppsec{2.2} and \supptab{2} finds
that the trained routing weights organize by wavelet subband
(NMI $0.243$ vs.\ $0.001$ for random routing) while carrying chance-level
linearly decodable degradation information.
Within each expert, the prompt-conditioned MLP (Eq.~\ref{eq:mag_mod})
produces channel-specific scale and bias that vary by task, so the same
expert applies different magnitude modulation to a rainy versus a noisy
input.
Empirically, routing organizes by \emph{what a subband looks like
spectrally}, while correction adapts to \emph{the task context encoded in
the prompt}.

\subsection{Fourier Expert}

Each Fourier expert receives a subband $\mathbf{s}$ and the pooled prompt
$\bar{\mathbf{p}}$.
We compute the 2-D real FFT, obtaining magnitude $m = |\mathbf{F}|$ and
phase $\varphi = \angle\mathbf{F}$.
A lightweight MLP $g_m: C_p \to 2C$ predicts per-channel scale $a$ and bias $b$,
and applies magnitude modulation:
\begin{equation}
  m' = \max\!\bigl(m \cdot \bigl(1 + \tanh(a)\bigr) + b,\; 0\bigr).
  \label{eq:mag_mod}
\end{equation}
The tanh-gated scale bounds the multiplicative factor to $(0,2)$; the
outer $\max(\cdot,0)$ clamp ensures $m'\!\ge\!0$ so that $\mathrm{polar}(m',\varphi)$
remains well-defined when the additive bias $b$ is negative.

A second, lighter MLP $g_\varphi: C_p \to C$ predicts a per-channel phase
scale $c$ and applies a bounded additive phase shift:
\begin{equation}
  \varphi' = \varphi + \tanh(c) \cdot \Delta_{\max}, \qquad \Delta_{\max} = \pi/4.
  \label{eq:phase_mod}
\end{equation}
The $\tanh$ gate bounds the phase perturbation to $\pm\pi/4$ ($45^\circ$),
preventing the additive shift from destructively overwriting the spatial
structure encoded in $\varphi$. The last layer of $g_\varphi$ is
zero-initialized, so $c\approx 0$ and $\varphi'\approx\varphi$ at the start
of training; the expert therefore begins as a near-identity map on phase
and only gradually learns a useful perturbation.

The modulated feature is reconstructed as $\hat{\mathbf{s}} =
\mathrm{irfft2}(\mathrm{polar}(m', \varphi'))$.
A lightweight spatial refinement (depthwise $3{\times}3$ + pointwise conv)
is applied after reconstruction.
Ablation A4 (Table~\ref{tab:ablation}) shows that removing phase
modulation (fixing $\varphi'=\varphi$) performs on par with the full
design ($+0.04$\,dB on average), indicating the zero-initialized phase
projector remains near identity throughout training.
All reported WaFIR results use the full configuration with phase modulation
(A3), consistently across text, figures, tables, and released code;
magnitude-only (A4) is an equally strong simplification.

\begin{figure}[t]
  \centering
  \includegraphics[width=\linewidth]{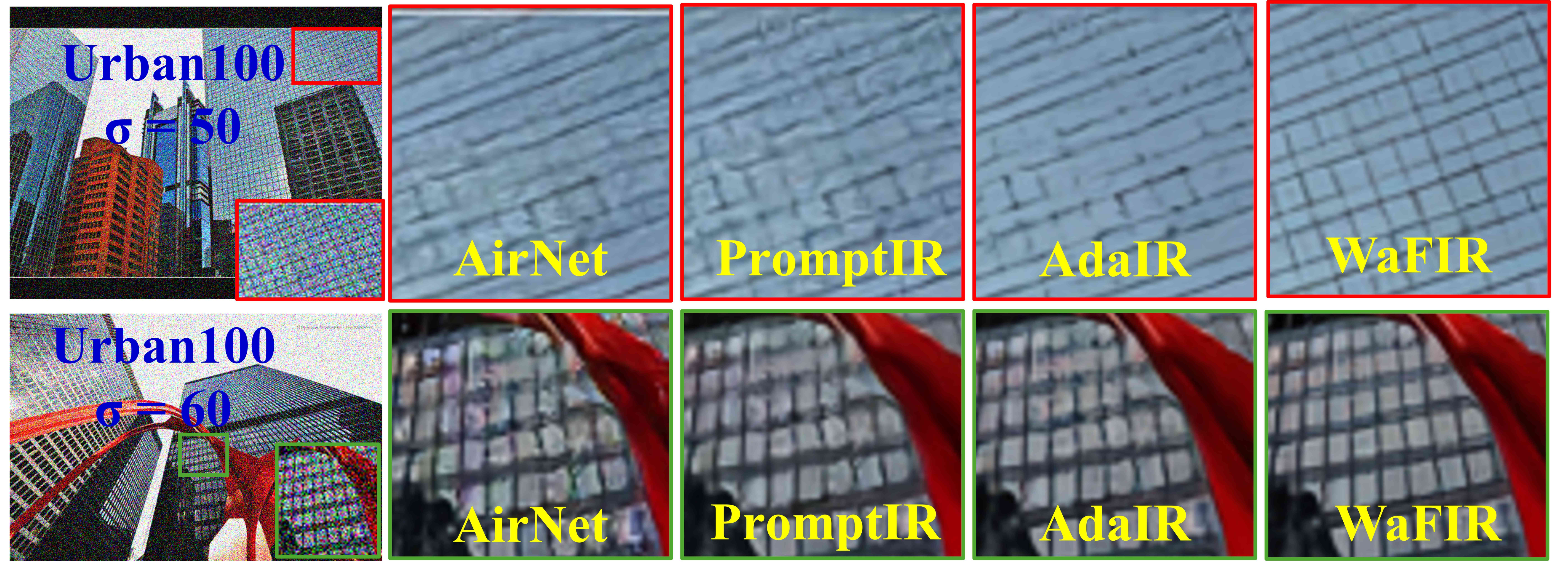}
  \caption{\textbf{Line-structure recovery on Urban100.} WaFIR restores thin,
  repeated lines coherently, without the ringing or block artifacts of spatial
  baselines, a direct benefit of subband-wise global Fourier modulation.}
  \label{fig:zoom_line_main}
\end{figure}

\subsection{Cross-Subband Interaction and Residual Gate}

After all four subbands are independently processed by their routed experts,
a grouped $1{\times}1$ convolution with $g{=}C$ groups enables cross-subband
interaction:
\begin{equation}
  \tilde{\mathbf{S}} = \mathrm{Conv}^{g=C}_{1\times1}
  ([\hat{\mathbf{s}}_\text{LL};\hat{\mathbf{s}}_\text{LH};
    \hat{\mathbf{s}}_\text{HL};\hat{\mathbf{s}}_\text{HH}]),
  \label{eq:cross}
\end{equation}
allowing, for example, the recovered illumination in LL to modulate texture
in HH. This costs only $4C$ parameters.
The spatial feature is reconstructed via IDWT: $\mathbf{r} = \mathrm{IDWT}(\tilde{\mathbf{S}})$.
The final block output uses a per-channel residual gate $\boldsymbol{\gamma}
\in \mathbb{R}^{C}$ initialized to zero, ensuring WaFIR behaves identically
to vanilla PromptIR at the start of training:
\begin{equation}
  \mathbf{x}' = \mathbf{x} + \boldsymbol{\gamma} \odot \mathrm{Conv}_{1\times1}(\mathbf{r}).
  \label{eq:residual}
\end{equation}

\subsection{Training Objective}

The total training loss combines pixel-wise $\ell_1$ reconstruction with an
auxiliary load-balance term:
\begin{equation}
  \mathcal{L} = \mathcal{L}_\text{rec} + \lambda\,\mathcal{L}_\text{lb},
  \quad \lambda = 0.01,
  \label{eq:loss}
\end{equation}
\begin{equation}
  \mathcal{L}_\text{lb} = \frac{1}{S}\sum_{s=1}^{S}\sum_{e=1}^{E}
  \Bigl(\bar{w}_{s,e} - \tfrac{1}{E}\Bigr)^2,
  \label{eq:lb}
\end{equation}
where $S{=}4$, $E{=}4$, and $\bar{w}_{s,e} = \frac{1}{B}\sum_b w^{(b)}_{s,e}$
is the mean routing weight for expert $e$ on subband $s$ over the batch.
The per-subband formulation discourages expert starvation within each
subband and stabilizes sparse routing. Its design rationale, comparison
with alternative balancing objectives, and sensitivity analysis are
provided in \suppsec{1.3} and \supptab{5}.

\begin{table*}[t]
\centering
\caption{%
  Quantitative comparison with state-of-the-art methods organized by
  architectural group. WaFIR uses the same standard training data as the
  baselines (PromptIR) on all three tasks, including the full RESIDE set for dehazing.
  Each cell: PSNR\,(dB)\,/\,SSIM.
  \textcolor{bestred}{Red}\,=\,best across the table for that task;
  \textcolor{secondblue}{Blue}\,=\,second best.
  ``--''\,=\,not publicly available.
}
\label{tab:sota_full}
\setlength{\tabcolsep}{4pt}
\resizebox{\textwidth}{!}{%
\begin{tabular}{l l c ccccc c}
\toprule
\multirow{2}{*}{\makecell[l]{Arch.\\Group}} &
\multirow{2}{*}{Method} &
\multirow{2}{*}{Venue} &
\multirow{2}{*}{\makecell{Dehazing\\SOTS~\cite{reside}}} &
\multirow{2}{*}{\makecell{Deraining\\Rain100L~\cite{rain100l}}} &
\multicolumn{3}{c}{Denoising (BSD68~\cite{bsd68})} &
\multirow{2}{*}{Average} \\
\cmidrule(lr){6-8}
 & & & & & $\sigma{=}15$ & $\sigma{=}25$ & $\sigma{=}50$ & \\
\midrule
\multirow{4}{*}{\makecell[l]{CNN-\\based}} &
  LPN\cite{lpn} & CVPR & 20.84/0.828 & 24.88/0.784 & 26.47/0.778 & 24.77/0.748 & 21.26/0.552 & 23.64/0.738 \\
& MPRNet\cite{mprnet} & CVPR & 28.00/0.958 & 33.86/0.958 & 33.27/0.920 & 30.67/0.871 & 27.29/0.761 & 30.63/0.894 \\
& ADFNet\cite{adfnet} & AAAI & 28.13/0.961 & 34.24/0.965 & 33.76/0.929 & 30.83/0.871 & 27.75/0.793 & 30.94/0.904 \\
& FSNet\cite{fsnet} & TPAMI & 29.14/0.969 & 35.61/0.969 & 33.81/0.874 & 30.84/0.872 & 27.69/0.762 & 31.42/0.906 \\
\midrule
\multirow{7}{*}{\makecell[l]{Trans-\\former-\\based}} &
  DehazeFormer\cite{dehazeformer} & TIP & 29.58/0.970 & 35.37/0.969 & 33.91/0.914 & 30.14/0.858 & 27.37/0.779 & 31.09/0.898 \\
& DRSformer\cite{drsformer} & CVPR & 29.02/0.968 & 35.89/0.970 & 33.28/0.921 & 30.55/0.862 & 27.58/0.786 & 31.26/0.902 \\
& Restormer\cite{restormer} & CVPR & 27.78/0.958 & 33.78/0.958 & 33.72/0.865 & 30.67/0.865 & 27.63/0.792 & 30.75/0.901 \\
& AirNet\cite{airnet} & CVPR & 27.94/0.962 & 34.90/0.967 & 33.92/0.933 & 31.26/0.888 & 28.00/0.797 & 31.20/0.910 \\
& IDR\cite{idr} & CVPR & 29.87/0.970 & 36.03/0.971 & 33.82/0.931 & 31.32/0.884 & 28.04/0.798 & 31.83/0.911 \\
& Gridformer\cite{gridformer} & IJCV & 30.37/0.970 & 37.15/0.972 & 33.93/0.931 & 31.37/0.887 & 28.11/0.801 & 32.19/0.912 \\
& NDR\cite{ndr_ir} & TIP & 28.64/0.962 & 35.42/0.969 & 34.04/0.931 & 31.36/0.887 & 28.10/0.798 & 31.51/0.910 \\
\midrule
\multirow{3}{*}{\makecell[l]{Visual\\Prompt-\\based}} &
  PromptIR\cite{promptir} & NeurIPS & 30.58/0.974 & 36.37/0.972 & 33.98/0.933 & 31.31/0.888 & 28.06/0.799 & 32.06/0.913 \\
& ProRes\cite{prores_ir} & arXiv & 28.38/0.938 & 33.68/0.954 & 32.10/0.907 & 30.18/0.863 & 27.58/0.779 & 30.38/0.888 \\
& Up-Restorer\cite{uprestorer} & AAAI & \textcolor{secondblue}{30.68}/\textcolor{secondblue}{0.977} & 36.74/0.978 & 33.99/0.933 & 31.33/0.888 & 28.07/0.799 & 32.16/0.915 \\
\midrule
\multirow{1}{*}{\makecell[l]{MoE}} &
  LoRA-IR\cite{lorair} & arXiv & \textcolor{secondblue}{30.68}/0.961 & 37.75/0.979 & 34.06/\textcolor{secondblue}{0.935} & 31.42/0.891 & 28.18/0.803 & 32.42/0.914 \\
\midrule
\multirow{1}{*}{\makecell[l]{SSM-based}} &
  MambaIR\cite{mambaIR} & ECCV & 29.57/0.970 & 35.42/0.969 & 33.88/0.917 & 30.95/0.874 & 27.74/0.793 & 31.51/0.907 \\
\midrule
\multirow{2}{*}{\makecell[l]{Freq. domain}} &
  FDGAN\cite{fdgan} & AAAI & 24.71/0.929 & 29.89/0.933 & 30.25/0.910 & 28.81/0.868 & 26.43/0.776 & 28.02/0.883 \\
& AdaIR\cite{adair} & ICLR & \textcolor{bestred}{31.06}/\textcolor{bestred}{0.980} & \textcolor{secondblue}{38.64}/\textcolor{secondblue}{0.983} & 34.12/\textcolor{secondblue}{0.935} & 31.45/\textcolor{secondblue}{0.892} & 28.19/0.802 & \textcolor{bestred}{32.69}/\textcolor{secondblue}{0.918} \\
\midrule
\multirow{2}{*}{\makecell[l]{VLM /\\Foundation}} &
  InstructIR-3D\cite{instructir3d} & ECCV & 30.22/0.959 & 37.98/{0.978} & \textcolor{secondblue}{34.15}/0.933 & \textcolor{secondblue}{31.52}/0.891 & \textcolor{bestred}{28.30}/\textcolor{secondblue}{0.804} & 32.43/0.913 \\
& DA-CLIP-SDE\cite{daclipsde} & ICLR & 26.83/0.962 & 35.85/0.972 & 31.43/0.889 & 25.05/0.605 & 18.33/0.308 & 27.50/0.747 \\
\midrule
\multicolumn{2}{l}{\cellcolor{wafirrow}\textbf{WaFIR (Ours)}} & \cellcolor{wafirrow}-- & \cellcolor{wafirrow}29.70/0.970 & \cellcolor{wafirrow}\textcolor{bestred}{38.99}/\textcolor{bestred}{0.985} & \cellcolor{wafirrow}\textcolor{bestred}{34.20}/\textcolor{bestred}{0.937} & \cellcolor{wafirrow}\textcolor{bestred}{31.54}/\textcolor{bestred}{0.894} & \cellcolor{wafirrow}\textcolor{secondblue}{28.29}/\textcolor{bestred}{0.807} & \cellcolor{wafirrow}\textcolor{secondblue}{32.54}/\textcolor{bestred}{0.919} \\
\bottomrule
\end{tabular}%
}
\end{table*}

\section{Experiments}
\label{sec:experiments}

\subsection{Implementation Details}
\label{sec:impl}

WaFIR is implemented in PyTorch and is trained end-to-end on all degradation
types simultaneously.
All experiments use the AdamW optimizer~\cite{loshchilov2017decoupled}
with an initial learning rate of $2 \times 10^{-4}$, $\beta_1 = 0.9$,
$\beta_2 = 0.999$, and weight decay $10^{-2}$.
The learning rate follows a linear warm-up over 15 epochs, then decays
to zero via cosine annealing over the remaining epochs.
We train for 150 epochs with a batch size of 32 and an input patch size
of $128 \times 128$ pixels, randomly cropped from training images.
Data augmentation consists of random horizontal and vertical flips.
Training uses BFloat16 mixed precision~\cite{kalamkar2019study} on 2 NVIDIA A100 GPUs.

The total loss combines pixel-wise $\ell_1$ reconstruction with the
auxiliary per-subband load-balance term (Eq.~\ref{eq:loss}),
with $\lambda{=}0.01$ fixed throughout all experiments.
WaFIR contains \textbf{42.15M} parameters in total,
of which 35.59M
belong to the PromptIR backbone and $+$6.56M are introduced by the
three SPMoE levels ($+$18.4\%).
Inference is evaluated at full image resolution on a single GPU
with no test-time augmentation.

\subsection{Datasets and Evaluation Metrics}
\label{sec:datasets}

We follow the standard all-in-one protocol of AirNet~\cite{airnet} and
PromptIR~\cite{promptir} across three degradation types.
For denoising, we train on BSD400~\cite{bsd400} and WED~\cite{ma2016waterloo}
(5{,}144 images) and test on BSD68~\cite{bsd68} and
Urban100~\cite{urban100} at $\sigma \in \{15, 25, 50\}$.
Urban100 serves as a cross-dataset generalization test: the AWGN process
matches training, but the urban-texture image domain is unseen.
For deraining, we train on Rain100L~\cite{rain100l} (200 pairs, repeat
factor 120 for multi-task balance~\cite{airnet}) and test on the
Rain100L test set (100 pairs).
For dehazing, we train on the full RESIDE dataset~\cite{reside} (72{,}135 training hazy images) and evaluate on the 500-image Synthetic Objective Testing Set (SOTS), following the same standard protocol as the baselines. Quantitative results are reported using PSNR, SSIM~\cite{wang2004image}, and LPIPS~\cite{zhang2018perceptual} on full RGB images without task-specific post-processing. We further provide qualitative and interpretability analyses using residual maps, Fourier spectral maps, and Local Attribution Maps (LAM)~\cite{gu2021interpreting} to evaluate reconstruction fidelity, frequency restoration, and the model's spatial attribution behavior.

\subsection{Results}
\label{sec:results}
\subsubsection{Comparisons with SOTA models:}

Table~\ref{tab:sota_full} compares WaFIR with 20 blind unified restoration methods under the same three-degradation protocol. WaFIR achieves the best PSNR on deraining (Rain100L, $38.99$\,dB) and on denoising at $\sigma{\in}\{15,25\}$ (and essentially ties the best at $\sigma{=}50$), together with the highest average SSIM ($0.919$). As illustrated in Figure~\ref{fig:qual_rain}, Figure~\ref{fig:zoom_line_main} and \suppfig{4}, these improvements are most pronounced for structured degradations such as rain and noise, whose directional or stationary spectral signatures can be effectively exploited by the proposed subband routing mechanism. Beyond the standard benchmark protocol, we further evaluate WaFIR under both single- and multiple-degradation settings. Detailed quantitative and qualitative analyses are provided in \suppsec{4}.

\subsubsection{Generalization under Distribution Shift:}

Without fine-tuning, we evaluate the same model on the CDD-11 rain test subset, Urban100 with both in-range and out-of-range noise levels ($\sigma=60$), and a synthetic denoising benchmark constructed from Scanning Acoustic Microscopy (SAM) images. These experiments evaluate dataset/domain shift and noise-intensity shift rather than an unseen degradation type. Table~\ref{tab:ood} reports PSNR, SSIM, and LPIPS on the in-range settings, where WaFIR leads on all three metrics; at the out-of-range $\sigma{=}60$ no single method dominates across all metrics (\supptab{7}). Qualitative cross-domain examples are shown in \suppfig{11}.

\begin{table}[t]
\centering
\setlength{\tabcolsep}{4pt}
\footnotesize
\caption{\textbf{Out-of-distribution generalization.} The all-in-one model
(trained on noise, haze, and rain) is tested \emph{directly} (no fine-tuning) on: an unseen
rain dataset (CDD-11), unseen natural-image denoising (Urban100), and an
\emph{industrial domain}, SAM, for electronic
semiconductor inspection (synthetic AWGN). Out-of-range $\sigma{=}60$ results
are reported in \supptab{7}. Metrics:
PSNR$\uparrow$/SSIM$\uparrow$/LPIPS$\downarrow$; best per block in \textbf{bold}.}
\label{tab:ood}
\begin{tabular}{@{}llccc@{}}
\toprule
Setting & Method & PSNR$\uparrow$ & SSIM$\uparrow$ & LPIPS$\downarrow$ \\
\midrule
\multirow{4}{*}{\shortstack[l]{CDD-11\\(rain)}}
 & PromptIR & 24.44 & 0.8638 & 0.1734 \\
 & AdaIR    & 24.73 & 0.8625 & 0.1691 \\
 & AirNet   & 23.07 & 0.8263 & 0.2198 \\
 & \textbf{WaFIR} & \textbf{24.94} & \textbf{0.8746} & \textbf{0.1577} \\
\midrule
\multirow{4}{*}{\shortstack[l]{Urban100\\avg\\($\sigma\!\in\!15\text{-}50$)}}
 & PromptIR & 30.77 & 0.9121 & 0.0612 \\
 & AdaIR    & 31.23 & 0.9193 & 0.0569 \\
 & AirNet   & 30.77 & 0.9135 & 0.0592 \\
 & \textbf{WaFIR} & \textbf{31.51} & \textbf{0.9249} & \textbf{0.0513} \\
\midrule
\multirow{4}{*}{\shortstack[l]{SAM\\avg\\($\sigma\!\in\!25\text{-}50$)}}
 & PromptIR & 31.42 & 0.8108 & 0.2859 \\
 & AdaIR    & 31.52 & 0.8128 & 0.2936 \\
 & AirNet   & 31.32 & 0.8049 & 0.3243 \\
 & \textbf{WaFIR} & \textbf{31.63} & \textbf{0.8152} & \textbf{0.2718} \\
\bottomrule
\end{tabular}
\end{table}

\begin{table}[t]
\centering
\caption{Ablation study. FULL $=$ mean PSNR (dB) over 8 conditions (BSD68 and
Urban100 at $\sigma{\in}\{15,25,50\}$, Rain100L, SOTS); $\Delta$ relative to A3.
\textcolor{green!60!black}{Green} $=$ gain, \textcolor{red!70!black}{red} $=$ drop.
Full per-task breakdown in \supptab{3}.}
\label{tab:ablation}
\setlength{\tabcolsep}{4pt}
\renewcommand{\arraystretch}{0.95}
\footnotesize
\begin{tabular}{llcc}
\toprule
ID & Component & FULL & $\Delta$ \\
\midrule
A3 & Full WaFIR (reference) & 30.35 & --- \\
\midrule
\multicolumn{4}{l}{\textit{Baseline}} \\
A0 & No SPMoE (PromptIR) & 29.87 & \textcolor{red!70!black}{$-$0.48} \\
\midrule
\multicolumn{4}{l}{\textit{SPMoE insertion depth}} \\
A1 & SPMoE at L3 only & 30.21 & \textcolor{red!70!black}{$-$0.14} \\
A2 & SPMoE at L2$+$L3 & 30.15 & \textcolor{red!70!black}{$-$0.20} \\
A15 & Encoder $+$ decoder SPMoE & 29.94 & \textcolor{red!70!black}{$-$0.41} \\
\midrule
\multicolumn{4}{l}{\textit{Routing quality}} \\
A5 & Random routing & 30.12 & \textcolor{red!70!black}{$-$0.23} \\
A6 & Top-$k{=}1$ routing & 30.00 & \textcolor{red!70!black}{$-$0.35} \\
\midrule
\multicolumn{4}{l}{\textit{Routing signal}} \\
A12 & Prompt-only router & 30.09 & \textcolor{red!70!black}{$-$0.26} \\
A14 & Descriptor-only router & 30.20 & \textcolor{red!70!black}{$-$0.15} \\
\midrule
\multicolumn{4}{l}{\textit{Frequency decomposition and interaction}} \\
A7 & No cross-subband & 30.19 & \textcolor{red!70!black}{$-$0.16} \\
A9 & Avg-pool DWT & 30.26 & \textcolor{red!70!black}{$-$0.10} \\
\midrule
\multicolumn{4}{l}{\textit{Expert design}} \\
A4 & No phase (magnitude only) & 30.39 & \textcolor{green!60!black}{$+$0.04} \\
\bottomrule
\end{tabular}
\end{table}

\subsection{Ablation Study}
\label{sec:ablation}

Table~\ref{tab:ablation} summarizes the principal ablations; \suppsec{3} provides the full interpretation and additional routing and regularization controls. All ablations use reduced training subsets ($1{,}200$ denoising, $200$ deraining, $391$ dehazing images) under the final model's protocol (architecture, optimizer, schedule, evaluation), so differences are attributable to the ablated component alone. Deltas are mean PSNR relative to the full model (A3).

\noindent\textbf{Decoder placement and three-level design (A1, A2, A15):}
Restricting SPMoE to L3 (A1, $-0.14$\,dB) or L2$+$L3 (A2, $-0.20$\,dB) underperforms the three-level design, and extending it to encoder levels degrades further (A15, $-0.41$\,dB). Spectral modulation helps specifically at decoder levels, where prompt context is injected and reconstruction occurs.

\noindent\textbf{Learned, soft routing (A5, A6):}
Random subband-to-expert assignment (A5, $-0.23$\,dB) confirms the router learns meaningful frequency-specialized patterns rather than arbitrary ones. Top-$k{=}1$ routing (A6, $-0.35$\,dB) gives the largest drop, showing that soft interpolation between experts lets each subband adapt to intermediate frequency characteristics (\suppfig{2}).

\noindent\textbf{Descriptor and prompt are complementary (A12--A14):}
Prompt-only routing (A12, $-0.26$\,dB) degrades performance and reorganizes routing around task identity rather than frequency. Descriptor-only routing (A14, $-0.15$\,dB) retains more performance than prompt-only routing in this controlled setting, showing that intrinsic spectral characteristics provide informative routing cues. The residual gap to joint conditioning reflects complementary prompt context, most visibly on deraining - degradation spectra overlap.

\noindent\textbf{Decomposition and cross-subband interaction (A7, A9):}
Removing the grouped $1{\times}1$ cross-subband convolution (A7, $-0.16$\,dB) shows inter-frequency interaction before IDWT aids coherent reconstruction, and replacing Haar DWT with average pooling (A9, $-0.10$\,dB) shows structured decomposition is needed for subband-specific descriptors.

\noindent\textbf{Phase modulation (A4):}
The bounded, zero-initialized phase branch converges near identity: removing it changes mean PSNR by only $+0.04$\,dB (within run-to-run variance), so magnitude modulation carries the restoration-relevant information. We keep the full expert for consistency with the released code; the magnitude-only variant is equivalent.

\noindent\textbf{Seed robustness:}
To confirm the ablation margins are not seed artifacts, we retrain both the full model (A3) and the no-SPMoE baseline (A0) with two random seeds (best-validation checkpoints; \supptab{6}). A3 attains FULL $=30.29{\pm}0.02$\,dB versus A0 $29.83{\pm}0.04$\,dB, and the $+0.46$\,dB SPMoE gain holds on both seeds ($+0.45$ and $+0.47$\,dB), far exceeding the seed spread. Because A3's seed deviation on FULL is only $0.02$\,dB, the component deltas of $\geq0.14$\,dB in Table~\ref{tab:ablation} all lie well outside seed noise.

\noindent\textbf{Routing under distribution shift (A3 vs.\ A13):}
The comparisons above are in-distribution, where a label-conditioned router (A13) receiving the ground-truth degradation label is marginally stronger ($+0.06$\,dB on average on BSD68). This edge inverts off-distribution (Table~\ref{tab:a13_main}; full results in \supptab{4}): A3 exceeds A13 by $+0.45$\,dB PSNR ($+0.046$ SSIM) at extrapolated $\sigma{=}70$ and by $+0.67$\,dB on rain$+$noise. On haze$+$noise the two are indistinguishable ($15.755$ vs.\ $15.739$\,dB), and both degrade sharply (roughly $12$--$18$\,dB below single-degradation performance), a shared failure mode rather than an advantage for either. In short, when the ground-truth label is unavailable or ill-defined, A3's continuous descriptor stays well-defined while A13 has no fallback, making label-free frequency routing the more robust choice off-distribution.

\begin{table}[t]
\centering
\footnotesize
\setlength{\tabcolsep}{4pt}
\caption{Learned (A3) vs.\ label-conditioned (A13) routing, PSNR (dB).
$\Delta{=}$A3$-$A13. A13's small in-distribution edge inverts off-distribution.
Full results in \supptab{4}.}
\label{tab:a13_main}
\begin{tabular}{lccc}
\toprule
Condition & A3 & A13 & $\Delta$ \\
\midrule
BSD68 $\sigma{=}25$ (in-dist.) & 30.96 & 31.02 & \textcolor{red!70!black}{$-$0.06} \\
$\sigma{=}35$ (interp.) & 29.31 & 29.27 & \textcolor{green!60!black}{$+$0.04} \\
$\sigma{=}70$ (extrap.) & 24.85 & 24.41 & \textcolor{green!60!black}{$+$0.45} \\
Rain$+$noise & 25.14 & 24.47 & \textcolor{green!60!black}{$+$0.67} \\
Haze$+$noise & 15.76 & 15.74 & \textcolor{green!60!black}{$+$0.02} \\
\bottomrule
\end{tabular}
\end{table}

\subsection{Analysis: Routing Self-Organization}
\label{sec:analysis}

\begin{figure}[t]
  \centering
  \includegraphics[width=\linewidth]{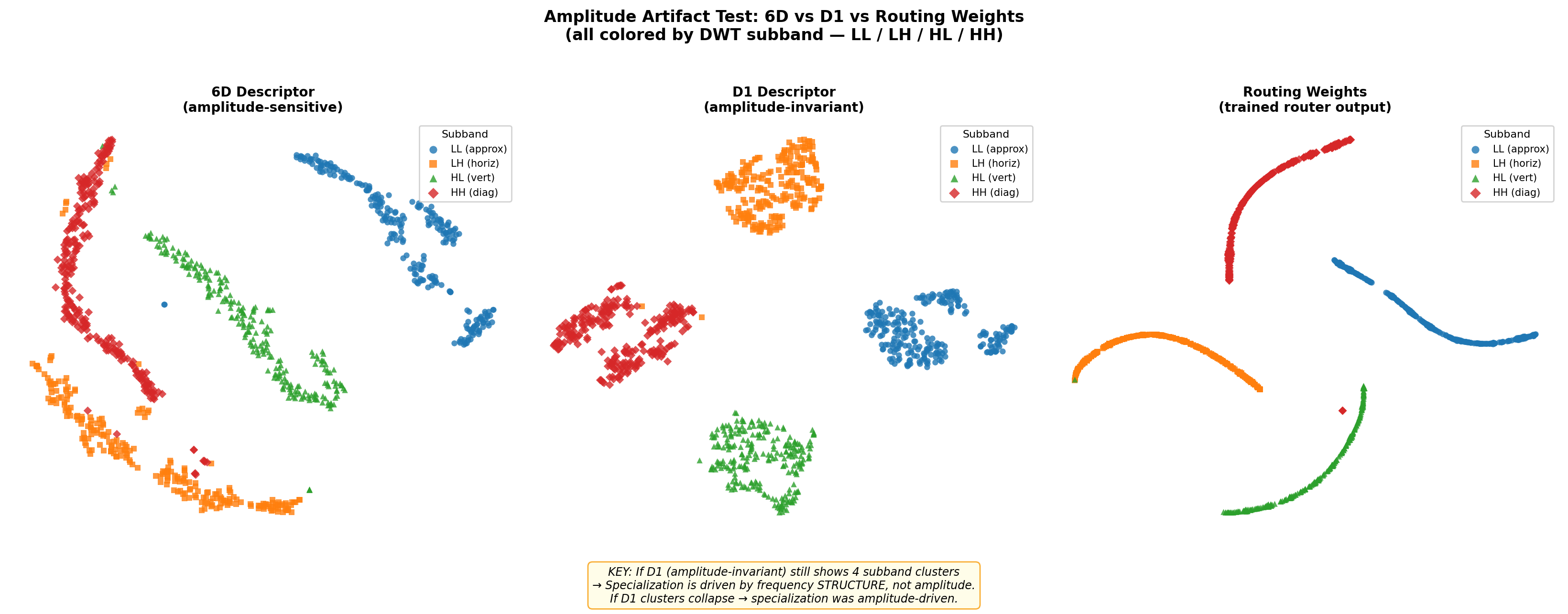}
  \caption{%
  \textbf{Amplitude artifact test.}
  t-SNE of three feature spaces, all colored by DWT subband.
  \emph{Left:} 6-D descriptor (amplitude-sensitive)---subbands separate but with scatter.
  \emph{Center:} D1 amplitude-invariant descriptor---four subbands form
  \textbf{tighter, cleaner clusters}, ruling out the amplitude-artifact hypothesis.
  \emph{Right:} Actual routing weight vectors of the trained router---four
  arcs consistent with D1 structure, confirming genuine frequency-band
  specialization.
  }
  \label{fig:amplitude_test}
\end{figure}

The main manuscript summarizes the central findings; the complete routing quantification, oracle-label control, and amplitude-invariance test are provided in \suppsec{2}.

\noindent\textbf{Routing self-organizes by frequency subspace:}
Projecting A3's routing weights with t-SNE and coloring by degradation type collapses all five conditions (Denoise$_{15/25/50}$, Derain, Dehaze) onto a single mixed manifold, whereas coloring the same vectors by DWT subband yields four distinct arcs without any subband-level supervision (\suppfig{2}). The organization is by frequency subspace, not degradation semantics: the LL subband of a rain image and of a haze image activate the same expert pair. Quantitatively ($N{=}250$), $k$-means NMI against the true subband label is $0.243$ for WaFIR, versus $0.001$ for random routing and $0.024$ for degradation labels; a linear degradation-class probe on routing weights reaches $35.1\%{\pm}1.9\%$ balanced accuracy, indistinguishable from the $33.3\%$ chance level (\supptab{2}).

\noindent\textbf{An apparent inductive bias toward structured degradations:}
The observed routing behavior is consistent with an inductive bias toward degradations that have a consistent, structured spectral signature, such as the directional streaks of rain and the stationary statistics of noise, where WaFIR's gains concentrate. Qualitatively, the same tendency appears in two representative cases, the artifact-free recovery of noise-corrupted line structure (Figure~\ref{fig:zoom_line_main}; \suppfig{6}) and of out-of-distribution rain streaks (\suppfig{10}), together with the wider attribution field of its Local Attribution Maps (\suppfig{7}).

\noindent\textbf{Ruling out amplitude artifacts:}
The four clusters could reflect amplitude differences intrinsic to the Haar DWT (LL carries more energy than HH) rather than learned behavior. We test this with an amplitude-normalized descriptor D1 built from energy ratios and the coefficient of variation:
\begin{equation}
  \mathbf{d}^{\text{D1}} = \Bigl[
    \log\bar{E},\;
    \tfrac{\sigma_E}{\bar{E}},\;
    \tfrac{E_\text{low}}{E_\text{tot}},\;
    \tfrac{E_\text{high}}{E_\text{tot}},\;
    \tfrac{E_h}{E_\text{tot}},\;
    \tfrac{E_v}{E_\text{tot}},\;
    \tfrac{E_\text{low}}{E_\text{high}}
  \Bigr],
  \label{eq:d1_invariant}
\end{equation}
where every dimension except $\log\bar{E}$ correlates weakly with image amplitude ($|r|{<}0.15$, versus $r{=}0.80$ for the strongest 6-D dimension). If the clusters were an amplitude artifact, removing amplitude should dissolve them; instead, the subbands form \emph{tighter} clusters under D1 than under the 6-D descriptor (Figure~\ref{fig:amplitude_test}). Subband structure is therefore genuinely distinct beyond amplitude, and the router exploits true frequency-spatial structure, consistent with the $-0.23$\,dB drop under random routing (A5).

\section{Conclusion}
\label{sec:conclusion}

We presented WaFIR, which inserts a Spectral Mixture-of-Experts (SPMoE)
module into each PromptIR decoder level to decompose features into Haar
subbands, summarize each with a compact 6-D Fourier descriptor, and route it
through a learned sparse gate to specialized Fourier experts.
Under a unified all-in-one protocol, WaFIR improves over PromptIR by
$+0.48$\,dB on average, and its routing self-organizes primarily by frequency
subspace rather than by degradation identity, as supported by t-SNE,
NMI/probe analysis, and an amplitude-invariance control.
The novelty lies in the \emph{granularity} and input-derived nature of the
routing signal, which complements the task context a prompt-based blind
restorer infers from the input. Moreover, our findings suggest that input-derived frequency priors and
generalization across datasets and restoration tasks provide a promising
foundation for medical and scientific imaging, and vision-language restoration
with large language models; further limitations are discussed in \suppsec{6}.

\section*{Acknowledgments}
This work was supported by the National Research Foundation of Korea (NRF) funded by the Korea government (MSIT)
under Grant 2022R1A5A8023404.
The authors gratefully acknowledge the resources provided by the University of Alabama at Birmingham IT-Research
Computing group for high performance computing (HPC) support and GPU time on the Cheaha supercomputer.
The authors also acknowledge support from the NSF ACCESS Award number CIS260853.

\bibliographystyle{aaai2027}
\bibliography{references}

\appendix
\twocolumn[{\begin{center}{\LARGE\bfseries Supplementary Material}\end{center}\vspace{1em}}]
\setcounter{section}{0}\setcounter{figure}{0}\setcounter{table}{0}\setcounter{equation}{0}\setcounter{algorithm}{0}
\renewcommand{\thesection}{S\arabic{section}}
\renewcommand{\thefigure}{S\arabic{figure}}
\renewcommand{\thetable}{S\arabic{table}}
\renewcommand{\theequation}{S\arabic{equation}}
\renewcommand{\thealgorithm}{S\arabic{algorithm}}
\definecolor{bestred}{RGB}{210,30,45}\definecolor{secondblue}{RGB}{35,90,180}
The central thesis of WaFIR is that its contribution lies in the
\emph{granularity} and \emph{nature} of the routing signal frequency-aware
expert routing performed at the level of individual wavelet subbands, which
complements the task context inferred by a prompt-based blind
restorer rather than merely in operating without an explicit degradation
label at test time. Label-free inference is a consequence of this design, not
its novelty; the novelty is that routing decisions are resolved per subband
from an input-derived spectral descriptor. This supplementary material
substantiates that thesis and expands the method, analysis, and experimental
claims summarized in the main manuscript. It is intended to be read together with the main paper: Section~S1 expands the Method section, Section~S2 supports the routing claims in the Introduction and Analysis section, Section~S3 expands the main ablation table, and Sections~S4--S5 provide the additional results and complexity evidence referenced from the Experiments section.
The organization follows the evaluation logic of the study.
Section~\ref{supp:method} gives the complete SPMoE forward pass and the
per-subband load-balance formulation.
Section~\ref{supp:routing} examines how routing self-organizes and includes
matched identity-conditioned controls.
Section~\ref{supp:ablation} reports component, routing, and regularization
ablations.
Section~\ref{supp:additional_results} provides additional qualitative,
and out-of-distribution evaluations.
Section~\ref{supp:complexity} analyzes parameter and computational overhead,
and Section~\ref{supp:limitations} discusses limitations and future work.

\section{Additional Method Details}
\label{supp:method}

\subsection{Subband Frequency Signatures}
\label{supp:subbands}

Table~\ref{tab:subbands} summarizes the frequency content of each Haar subband
and the degradation type it primarily captures, as referenced from the Method
section of the main paper.

\begin{table}[h]
\centering
\caption{Frequency content and primary degradation signal per Haar subband.}
\label{tab:subbands}
\resizebox{\linewidth}{!}{%
\begin{tabular}{lll}
\toprule
Subband & Frequency content & Primary degradation \\
\midrule
LL & Low-low (approximation) & Haze (DC$\uparrow$) \\
LH & Low-high (horiz.\ edges) & Rain streaks (vert.\ spatial $\to$ horiz.\ freq.) \\
HL & High-low (vert.\ edges)  & Directional patterns, vertical streaks \\
HH & High-high (diagonal)     & Additive noise (flat high-freq.\ spectrum) \\
\bottomrule
\end{tabular}}
\end{table}

\subsection{Complete SPMoE Forward Pass}
\label{supp:algo}

Algorithms~\ref{alg:spmoe_block}--\ref{alg:spmoe_sub} give the complete
SPMoE block, including the spectral descriptor, router, and Fourier spectral-modulation expert.

\begin{algorithm}[t]
\caption{SpectralMoE Block (SPMoEBlock) --- full forward pass}
\label{alg:spmoe_block}
\begin{algorithmic}[1]
\Require Feature map $\mathbf{x}\in\mathbb{R}^{B\times C\times H\times W}$,
prompt $\mathbf{p}\in\mathbb{R}^{B\times C_p\times H_p\times W_p}$,
$E=4$ experts per subband, top-$k=2$ routing
\Ensure Refined feature $\hat{\mathbf{x}}\in\mathbb{R}^{B\times C\times H\times W}$

\Statex
\Statex \textit{--- Wavelet decomposition ---}
\State $\bar{\mathbf{x}} \gets \mathrm{ReflectPad}(\mathbf{x})$
\Comment{pad $H,W$ to even dimensions}
\State $\mathbf{s}_{LL},\mathbf{s}_{LH},\mathbf{s}_{HL},\mathbf{s}_{HH}
\gets \mathrm{HaarDWT}(\bar{\mathbf{x}})$
\Comment{each $\in\mathbb{R}^{B\times C\times H/2\times W/2}$}

\Statex
\Statex \textit{--- Per-subband routing and aggregation ---}
\For{$\mathbf{s} \in \{\mathbf{s}_{LL},\mathbf{s}_{LH},\mathbf{s}_{HL},\mathbf{s}_{HH}\}$}
    \State $\mathbf{d} \gets \Call{SpectralDescriptor}{\mathbf{s}}$
    \Comment{$\mathbf{d}\in\mathbb{R}^{B\times 6}$}
    \State $\mathbf{w} \gets \Call{SpectralRouter}{\mathbf{d},\mathbf{p}}$
    \Comment{$\mathbf{w}\in\mathbb{R}^{B\times E}$}
    \State $\hat{\mathbf{s}} \gets \sum_{e=1}^{E} w_e\,
    \mathrm{FourierExpert}_e(\mathbf{s},\mathbf{p})$
\EndFor

\Statex
\Statex \textit{--- Cross-subband interaction ---}
\State $\mathbf{Z} \gets
[\hat{\mathbf{s}}_{LL};\hat{\mathbf{s}}_{LH};
\hat{\mathbf{s}}_{HL};\hat{\mathbf{s}}_{HH}]$
\Comment{$\mathbf{Z}\in\mathbb{R}^{B\times 4C\times H/2\times W/2}$}
\State $\mathbf{Z} \gets \mathrm{GroupedConv}_{1\times1}(\mathbf{Z})$

\Statex
\Statex \textit{--- Reconstruction and residual connection ---}
\State $\mathbf{r} \gets \mathrm{HaarIDWT}(\mathbf{Z})_{:,:,:H,:W}$
\State $\hat{\mathbf{x}} \gets \mathbf{x}
+ \boldsymbol{\gamma}\odot \mathrm{Conv}_{1\times1}(\mathbf{r})$
\Comment{$\boldsymbol{\gamma}\in\mathbb{R}^{C}$, zero-initialized}
\State \Return $\hat{\mathbf{x}}$

\end{algorithmic}
\end{algorithm}

\begin{algorithm}[t]
\caption{Sub-procedures used in Algorithm~\ref{alg:spmoe_block}}
\label{alg:spmoe_sub}
\begin{algorithmic}[1]

\Function{SpectralDescriptor}{$\mathbf{s}$}
    \State $\mathbf{M} \gets |\mathrm{rFFT2}(\mathbf{s})|$
    \State $E_{\mathrm{low}} \gets
    \mu(\mathbf{M}_{:,:,0:l_H,0:l_W})$
    \State $E_{\mathrm{high}} \gets
    \mu(\mathbf{M}_{:,:,l_H:,l_W:})$
    \State $E_h \gets
    \mu(\mathbf{M}_{:,:,H'/4:H'/2,:})$
    \State $E_v \gets
    \mu(\mathbf{M}_{:,:,:,W'/4:})$
    \State \Return $[\mu(\mathbf{M}),\mathrm{Var}(\mathbf{M}),
    E_{\mathrm{low}},E_{\mathrm{high}},E_h,E_v]$
\EndFunction

\Statex

\Function{SpectralRouter}{$\mathbf{d},\mathbf{p}$}
    \State $\bar{\mathbf{p}} \gets \mathrm{AvgPool}(\mathbf{p}).\mathrm{flatten}()$
    \State $\mathbf{g} \gets
    W_2\,\mathrm{GELU}(W_1[\mathbf{d};\bar{\mathbf{p}}])/\tau$
    \State Keep only top-$k$ entries of $\mathbf{g}$ and set others to $-\infty$
    \State \Return $\mathrm{softmax}(\mathbf{g})$
\EndFunction

\Statex

\Function{FourierExpert$_e$}{$\mathbf{s},\mathbf{p}$}
    \State $\bar{\mathbf{p}} \gets \mathrm{AvgPool}(\mathbf{p}).\mathrm{flatten}()$
    \State $[a,b] \gets
    \mathrm{split}(\mathrm{MLP}_{\mathrm{mag}}(\bar{\mathbf{p}}),C)$
    \State $\mathbf{s}_{\varphi} \gets
    \mathrm{MLP}_{\mathrm{phase}}(\bar{\mathbf{p}})$
    \State $\mathcal{F} \gets \mathrm{rFFT2}(\mathbf{s})$
    \State $M \gets |\mathcal{F}|,\quad \varphi \gets \angle\mathcal{F}$
    \State $M' \gets
    \max\!\bigl(M(1+\tanh(a))+b,\,0\bigr)$
    \State $\varphi' \gets
    \varphi+\tanh(\mathbf{s}_{\varphi})\cdot\frac{\pi}{4}$
    \State $\mathbf{r} \gets
    \mathrm{irFFT2}(\mathrm{polar}(M',\varphi'))$
    \State \Return $\mathbf{r}+
    \mathrm{PWConv}(\mathrm{DWConv}(\mathbf{r}))$
\EndFunction

\end{algorithmic}
\end{algorithm}

\subsection{Per-Subband Load-Balance Regularization}
\label{supp:lb}

Sparse spectral routing is only useful if the four experts remain distinct.
This section explains why a load-balance term is required, why we apply it
\emph{per subband}, and quantifies its effect.

\subsubsection{Why Regularization Is Required}

Sparse top-$k$ routing is prone to \emph{expert collapse}: without
regularization the router quickly favors one or two experts for all inputs,
so the rest receive no gradient and remain uninitialized
\citep{switch_transformer,gshard}. In WaFIR this is fatal, because the entire
rationale for four experts is that each specializes in a distinct frequency
subband. Under collapse, all subbands route to the same pair and SPMoE
reduces to a dense layer with extra parameters and no specialization.

\subsubsection{Formulation}

We use the same mean-squared imbalance penalty introduced in the main manuscript, with an explicit uniform reference target:
\begin{equation}
  \mathcal{L}_\text{lb}
  = \frac{1}{S}\sum_{s=1}^{S}\sum_{e=1}^{E}
  \Bigl(\bar{w}_{s,e} - \tfrac{1}{E}\Bigr)^2 .
\end{equation}
Relative to the Switch auxiliary loss
$\alpha\sum_e f_e P_e$~\citep{switch_transformer}, which multiplies two
router-dependent quantities and sets no explicit target, and to entropy
maximization $-\sum_e \bar{w}_e\log\bar{w}_e$, whose gradient vanishes near
uniformity precisely when regularization is most needed,
$\mathcal{L}_\text{lb}$ has three practical properties: it encodes an
explicit target ($1/E=0.25$); its quadratic penalty grows with the severity
of imbalance and relaxes as balance is restored; and it is differentiable
and numerically stable everywhere.

\subsubsection{Why Per-Subband, Not Global}

A \emph{global} penalty $\sum_e(\bar{w}_e-1/E)^2$, with $\bar{w}_e$ averaged
over all subbands, admits a degenerate optimum: the router can send
$\text{LL}\to\{E_1,E_2\}$ and $\text{LH}\to\{E_3,E_4\}$ exclusively routing
each subband to a fixed pair while appearing globally balanced. More
generally, any permutation assigning a fixed subset of subbands to a fixed
pair satisfies a global constraint while providing no per-subband
specialization. The \emph{per-subband} form removes this global loophole by penalizing
persistent expert starvation separately within each subband. It does not
force the final router to remain uniform; rather, it keeps all experts
trainable early in optimization while allowing nonuniform, subband-dependent
specialization to emerge as the reconstruction objective becomes dominant.

\section{Routing Analysis}
\label{supp:routing}

\subsection{Frequency-Domain Motivation and Routing Self-Organization}

\begin{figure*}[t]
  \centering
  \includegraphics[width=\linewidth]{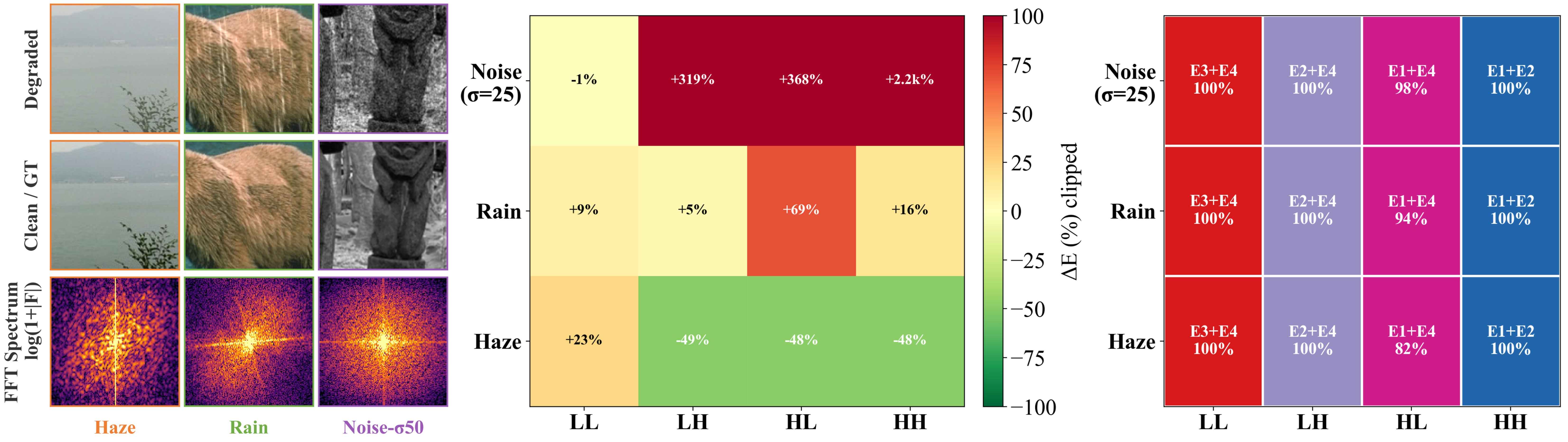}
  \caption{%
    \textbf{Frequency-domain motivation for WaFIR.}
    \emph{Left:} log-magnitude FFT spectra ($\log(1{+}|\mathcal{F}|)$) for
    three representative degradations. Haze concentrates energy in the DC
    component with attenuated high frequencies; rain produces directional
    stripe patterns aligned with streak orientation; Gaussian noise
    ($\sigma{=}50$) spreads energy across all bands.
    \emph{Middle:} relative subband energy change (\%) versus clean images
    across Haar subbands (LL, LH, HL, HH), clipped at $\pm100\%$ for
    display (Noise-HH actual: $+$2{,}214\%). Each degradation corrupts a
    distinct subband: noise inflates all high-frequency bands; haze raises
    LL by $+$23\% while suppressing LH/HL/HH by ${\approx}48\%$; rain
    concentrates energy in HL ($+$69\%).
    \emph{Right:} most frequent top-$k$ expert pair selected per
    (degradation, subband), averaged over 50 test images. Each subband is
    routed to a distinct pair (LL${\to}$E3+E4; LH${\to}$E2+E4;
    HL${\to}$E1+E4; HH${\to}$E1+E2) with 82--100\% consistency regardless
    of degradation type, indicating that routing self-organizes by
    frequency subspace without degradation-label or subband-level
    supervision.
  }
  \label{fig:motivation}
\end{figure*}

\begin{figure}[t]
  \centering
  \includegraphics[width=\linewidth]{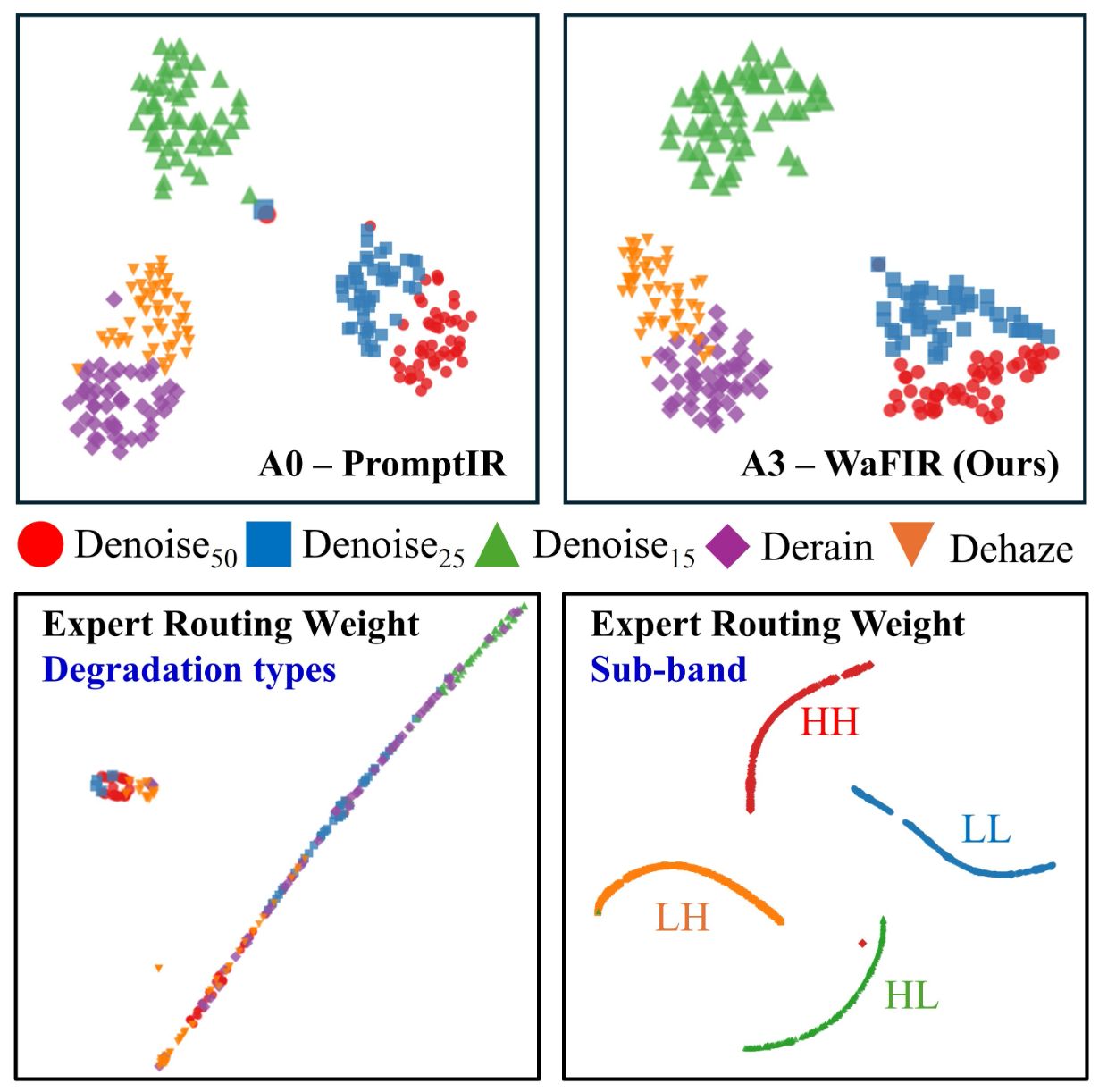}
  \caption{%
  \textbf{t-SNE of bottleneck features and routing weights.}
  \emph{Top:} bottleneck features of A0 (PromptIR, left) and A3 (WaFIR,
  right), colored by degradation type; WaFIR yields clearer inter-class
  separation, notably between Denoise$_{15}$/Denoise$_{25}$ and
  Derain/Dehaze.
  \emph{Bottom-left:} A3 routing-weight vectors colored by degradation
  type - all types lie on one mixed manifold, so routing does \emph{not}
  organize by degradation identity.
  \emph{Bottom-right:} the same vectors colored by DWT subband - four
  distinct arcs emerge without subband supervision, so routing organizes by
  frequency subspace. Routing is thus governed by \emph{which subband} a
  feature belongs to, not \emph{which degradation} produced it.
  }
  \label{fig:tsne}
\end{figure}

We first recall the qualitative evidence
(Figures~\ref{fig:motivation}--\ref{fig:tsne}), then test the claim with
matched controls (Table~\ref{tab:routing-quant}), and finally report a
controlled oracle-routing experiment (A13) that isolates the effect of the
organizing principle itself.

\subsection{Quantitative Routing Organization}
\label{supp:routing_quant}

The t-SNE plots are suggestive but qualitative. We test the claim by asking
two \emph{separate} questions about the trained router's per-subband weight
vectors, evaluated on a held-out set spanning all degradation types
($N{=}250$; 50 per degradation, 5 conditions): (i)~do the vectors cluster by
wavelet subband? and (ii)~do they encode degradation identity?
Every quantity is reported against a matched control, since an absolute
value is otherwise uninterpretable; permutation floors
(${\approx}\,0.01$ for all MI values) are estimated by shuffling labels
$100\times$.

\paragraph{Routing clusters by wavelet subband.}
$k$-means ($k{=}4$) on the routing-weight vectors, scored by Normalized
Mutual Information (NMI) against the true subband label, gives a clear
ordering (Table~\ref{tab:routing-quant}). Random routing yields
NMI$\,{=}\,0.001$ (at the floor). The 6-D spectral descriptor \emph{before}
the router MLP attains NMI$\,{=}\,0.166$; some subband structure is expected
here by construction, since the descriptor is derived from per-subband
Fourier statistics. The trained router reaches NMI$\,{=}\,0.243$, a clear
increment over the descriptor-only baseline ($0.166\!\to\!0.243$),
indicating that the MLP \emph{amplifies} subband-aligned structure rather
than merely inheriting it.

\paragraph{Degradation information concentrates in physically grounded subbands.}
We compute the mutual information between routing weights and the
degradation label \emph{per subband} (\texttt{mutual\_info\_classif},
$k$-NN based). Mean Deg-MI is $0.091$ (floor ${\approx}\,0.01$) and is
unevenly distributed: it concentrates in LL ($0.20$) and HL ($0.09$), while
LH ($0.04$) and HH ($0.03$) sit near the floor. At LL, all three
degradation pairs are comparably separable ($0.14$--$0.17$), consistent with
LL retaining global low-frequency statistics that every degradation
modifies. At HL, separability is driven specifically by rain (noise--rain
$0.09$, rain--haze $0.13$; noise--haze $0.017$), consistent with the
directional signature of rain streaks in the vertical-edge subband.
Degradation therefore surfaces in routing \emph{precisely where} the
corresponding wavelet subband physically encodes degradation-discriminative
information---a stronger statement than ``routing ignores degradation.''

\paragraph{Routing weights carry little linearly-decodable degradation information.}
A logistic-regression probe (5-fold CV, balanced accuracy) on the spectral
descriptor recovers degradation type at $40.5\%$, above the $33.3\%$
three-class chance level. On the routing weights the same probe yields
$35.1\%\!\pm\!1.9\%$, statistically indistinguishable from chance. The MLP
transforms away most linearly-accessible degradation structure; the residual
Deg-MI in LL and HL reflects nonlinear dependence.

\subsection{Controlled Identity-vs-Frequency Test}
\label{supp:oracle}

\paragraph{Oracle label-conditioned control (A13).}
The analyses above characterize what WaFIR \emph{does}; A13 tests whether the
frequency-based principle is \emph{preferable} to an identity-based one under
matched capacity. A13 is architecturally identical to WaFIR (A3) but replaces
the spectral descriptor with a routing signal derived from the \emph{oracle
degradation label}: every subband of an image is routed by the same one-hot
label. As expected, A13's routing weights carry far stronger degradation
information (Deg-MI $=0.519$ vs.\ $0.091$), uniformly distributed across
subbands (LL/LH/HL/HH $\approx0.50$--$0.57$), and subband organization
collapses (Sub-NMI $=0.024$ vs.\ $0.243$). The two models thus operate under
genuinely different principles.

Crucially, oracle labels do \emph{not} translate into uniformly better
restoration. Under the controlled reduced-budget evaluation
(Table~\ref{tab:ood_routing}), A13's in-distribution advantage is marginal
($+0.06$\,dB on average on BSD68) and reverses under distribution shift, where
A3's continuous descriptor stays well-defined while a one-hot label has no
fallback. WaFIR attains this without any degradation classifier, whereas A13
requires ground-truth labels that are unavailable in deployment. The mechanism
is that the descriptor encodes noise energy continuously
(\texttt{var\_mag}$\,{\propto}\,\sigma^2$), enabling per-image adaptation
\emph{within} a noise level that a one-hot label cannot represent.

\begin{table*}[h]
\centering
\small
\setlength{\tabcolsep}{6pt}
\caption{%
  Routing quantification ($N{=}250$ images, 5 degradation conditions,
  3 decoder levels, 4 subbands).
  \textbf{Sub-NMI}: NMI between $k$-means clusters of routing weights and the
  true subband label (higher $=$ more subband-organized).
  \textbf{Deg-MI}: MI between routing weights and degradation label, mean
  over subbands with per-subband breakdown (lower $=$ less
  degradation-organized).
  \textbf{Probe}: balanced accuracy of a linear degradation classifier
  (three-class chance $=33.3\%$). Permutation floor ${\approx}\,0.01$ for all
  MI values. WaFIR (A3) is strongly subband-organized and weakly
  degradation-organized; the oracle (A13) is the reverse.
}
\label{tab:routing-quant}
\begin{tabular}{lccccccc}
\toprule
Variant & Sub-NMI & Deg-MI & LL & HL & LH & HH & Probe \\
\midrule
Random (A5)          & 0.001 & 0.006 & 0.00 & 0.01 & 0.01 & 0.00 & --- \\
Descriptor (pre-MLP) & 0.166 & ---   & ---  & ---  & ---  & ---  & 40.5\% \\
\textbf{WaFIR (A3)}  & \textbf{0.243} & 0.091 & 0.20 & 0.09 & 0.04 & 0.03 & $35.1\%{\pm}1.9\%$ \\
A13 (oracle)         & 0.024 & 0.519 & 0.50 & 0.49 & 0.57 & 0.52 & $65.0\%{\pm}2.1\%$ \\
\bottomrule
\end{tabular}
\end{table*}

\subsection{Amplitude-Invariance Control}
\label{supp:amplitude}

To test whether the observed subband clusters are a trivial consequence of
absolute amplitude differences, we compare the original descriptor with an
amplitude-invariant ratio descriptor (D1) and with the trained routing weights.

\begin{figure*}[t]
  \centering
  \includegraphics[width=\linewidth]{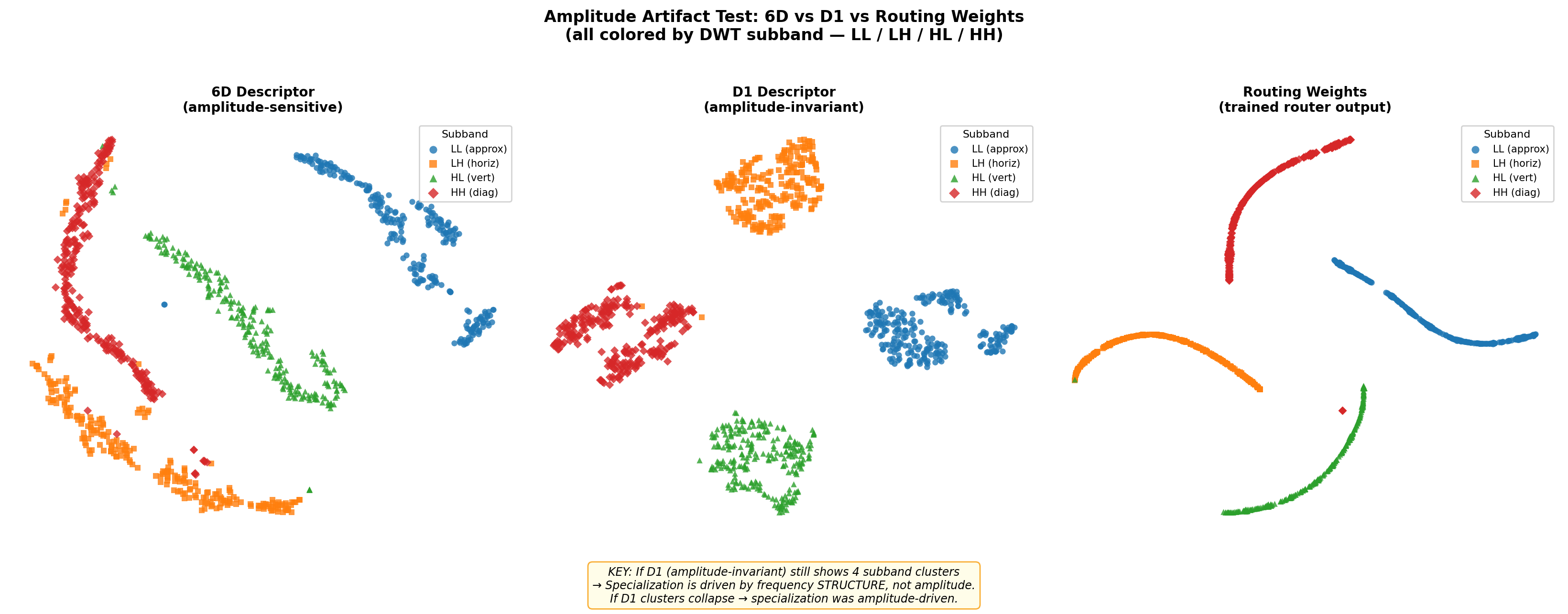}
  \caption{%
  \textbf{Amplitude artifact test.}
  t-SNE of three feature spaces, all colored by DWT subband.
  \emph{Left:} 6-D descriptor (amplitude-sensitive)---subbands separate but with scatter.
  \emph{Center:} D1 amplitude-invariant descriptor---four subbands form
  \textbf{tighter, cleaner clusters}, ruling out the amplitude-artifact hypothesis.
  \emph{Right:} Actual routing weight vectors of the trained router---four
  arcs consistent with D1 structure, confirming genuine frequency-band
  specialization.
  }
  \label{sfig:amplitude_test}
\end{figure*}

\section{Ablation Studies}
\label{supp:ablation}

This section expands the ablation table reported in the main manuscript. For completeness, we reproduce the controlled component table and then report analyses that do not fit in the main paper, including the label-conditioned routing test and the load-balance sensitivity study.

\subsection{Component-Level Ablation}
\label{supp:component_ablation}

Table~\ref{stab:ablation} reports the component-level ablation. Each row replaces
or removes a single element of the full model (A3) while holding the
architecture, optimizer, schedule, and evaluation fixed, so the resulting change
in the FULL metric---the mean PSNR over the eight test conditions (BSD68 and
Urban100 at $\sigma{\in}\{15,25,50\}$, Rain100L, and SOTS)---isolates that
element's contribution. The variants are grouped by the design question they
probe: the no-SPMoE \emph{baseline} (A0); SPMoE \emph{insertion depth}
(A1, A2, A15); \emph{routing quality} (random and top-$1$ routing, A5, A6); the
\emph{routing signal} that conditions the gate (A12, A14); \emph{frequency
decomposition and interaction} (A7, A9); and \emph{expert design} (A4). The
$\Delta$ column is the FULL difference relative to A3
(\textcolor{green!60!black}{green} $=$ gain,
\textcolor{red!70!black}{red} $=$ drop). All variants are trained on the reduced
subsets described below, so their absolute PSNR values are comparable to one
another but not to the full-data results in the main paper (Table~1); the
seed-stability analysis (Table~\ref{tab:seed}) confirms that deltas of
$\geq0.14$\,dB exceed run-to-run variation. Each group is interpreted in
Section~\ref{supp:ablation_discussion}.

\begin{table*}[t]
\centering
\caption{%
Ablation study.   {FULL} = mean PSNR (dB) over 8 test conditions:
BSD68 $\sigma{\in}\{15,25,50\}$, Urban100 $\sigma{\in}\{15,25,50\}$,
Rain100L, SOTS.
$\Delta$ relative to A3.
\textcolor{green!60!black}{Green} $\Delta$ = gain, \textcolor{red!70!black}{red} = drop.
}
\label{stab:ablation}
\setlength{\tabcolsep}{3.5pt}
\resizebox{\textwidth}{!}{%
\begin{tabular}{l l ccc ccc cc r r}
\toprule
\multirow{2}{*}{ID} & \multirow{2}{*}{Component} &
\multicolumn{3}{c}{BSD68 (PSNR)} &
\multicolumn{3}{c}{Urban100 (PSNR)} &
\multirow{2}{*}{\makecell{Rain\\100L}} &
\multirow{2}{*}{SOTS} &
\multirow{2}{*}{FULL} &
\multirow{2}{*}{$\Delta$} \\
\cmidrule(lr){3-5}\cmidrule(lr){6-8}
& & $\sigma{=}15$ & $\sigma{=}25$ & $\sigma{=}50$
  & $\sigma{=}15$ & $\sigma{=}25$ & $\sigma{=}50$ & & & & \\
\midrule
A3 &   {Full WaFIR} (reference)
   & 33.60 & 31.00 & 27.78 & 32.71 & 30.28 & 26.75 & 33.14 & 27.56
   & 30.35 & --- \\
\midrule
\multicolumn{12}{l}{\textit{Baseline}} \\
A0 & No SPMoE (PromptIR)
   & 33.45 & 30.87 & 27.63 & 32.31 & 29.77 & 26.34 & 31.57 & 27.06
   & 29.87 & \textcolor{red!70!black}{$-$0.48} \\
\midrule
\multicolumn{12}{l}{\textit{SPMoE insertion depth}} \\
A1 & SPMoE at L3 only
   & 33.59 & 30.99 & 27.73 & 32.81 & 30.19 & 26.63 & 32.33 & 27.42
   & 30.21 & \textcolor{red!70!black}{$-$0.14} \\
A2 & SPMoE at L2$+$L3
   & 33.51 & 30.94 & 27.71 & 32.80 & 30.17 & 26.59 & 32.11 & 27.39
   & 30.15 & \textcolor{red!70!black}{$-$0.20} \\
A15& Encoder $+$ decoder SPMoE
   & 33.49 & 30.87 & 27.60 & 32.53 & 29.95 & 26.42 & 31.56 & 27.12
   & 29.94 & \textcolor{red!70!black}{$-$0.41} \\
\midrule
\multicolumn{12}{l}{\textit{Routing quality}} \\
A5 & Random routing
   & 33.58 & 30.95 & 27.68 & 32.78 & 30.17 & 26.62 & 32.28 & 26.93
   & 30.12 & \textcolor{red!70!black}{$-$0.23} \\
A6 & Top-$k{=}1$ routing
   & 33.49 & 30.90 & 27.65 & 32.70 & 30.09 & 26.52 & 32.05 & 26.60
   & 30.00 & \textcolor{red!70!black}{$-$0.35} \\
\midrule
\multicolumn{12}{l}{\textit{Routing signal}} \\
A12& Prompt-only router (no spectral descriptor)
   & 33.47 & 30.91 & 27.68 & 32.51 & 29.95 & 26.47 & 32.16 & 27.55
   & 30.09 & \textcolor{red!70!black}{$-$0.26} \\
A14& Descriptor-only router
   & 33.54 & 30.95 & 27.73 & 32.44 & 29.93 & 26.53 & 32.73 & 27.73
   & 30.20 & \textcolor{red!70!black}{$-$0.15} \\
\midrule
\multicolumn{12}{l}{\textit{Frequency decomposition and interaction}} \\
A7 & No cross-subband interaction
   & 33.54 & 30.94 & 27.69 & 32.70 & 30.16 & 26.61 & 32.63 & 27.26
   & 30.19 & \textcolor{red!70!black}{$-$0.16} \\
A9 & Avg-pool DWT (no Haar wavelet)
   &   {33.62} & 31.03 & 27.78 & 32.71 & 30.19 & 26.75 & 32.55 & 27.41
   & 30.26 & \textcolor{red!70!black}{$-$0.10} \\
\midrule
\multicolumn{12}{l}{\textit{Expert design}} \\
A4 & No phase modulation (magnitude only)
   & 33.59 & 31.02 & 27.78 & 32.85 & 30.25 & 26.72 & 33.26 & 27.65
   & 30.39 & \textcolor{green!60!black}{$+$0.04} \\
\bottomrule
\end{tabular}}
\end{table*}

\subsection{Interpretation of Component Ablations}
\label{supp:ablation_discussion}

In this section, we conduct controlled ablation studies to evaluate the contribution of each component in WaFIR. All ablation variants are trained on reduced subsets of the original datasets (1,200 images for denoising, 200 images for deraining, and 391 images for dehazing) while preserving the network architecture outside the modified component, optimization settings, training schedule, and evaluation protocol. This matched setting is designed to attribute performance differences to the ablated component rather than to changes in data or optimization.

\noindent\textbf{Both decoder placement and three-level design are critical (A1, A2, A15).}
Adding SPMoE at L3 only (A1, $\Delta{=}{-}0.14$) or at L2$+$L3 (A2, $\Delta{=}{-}0.20$\,dB)
falls significantly short of the full three-level configuration.
We attribute this to the complementary nature of multi-scale frequency
correction: the coarsest block (L3, $C{=}384$) modulates global spectral
shape, while shallower blocks (L2, L1) address scale-specific textures.
Extending SPMoE to encoder levels as well (A15, $\Delta{=}{-}0.41$\,dB)
actively degrades performance, indicating that spectral expert modulation
is beneficial specifically at decoder levels---where prompt context is
injected and reconstruction occurs---rather than throughout the
network.

\noindent\textbf{Learned routing and soft interpolation are both essential (A5, A6).}
Random subband-to-expert assignment (A5, $\Delta=-0.23$\,dB) demonstrates that the learned router captures meaningful frequency-specialized routing patterns rather than arbitrary assignments. Restricting the router to top-$k=1$ (A6, $\Delta=-0.35$\,dB) results in the largest performance drop, highlighting the importance of soft interpolation between multiple experts. As illustrated by the arc geometry in Figure~\ref{fig:tsne}, weighted combinations of two experts allow each subband to adapt continuously to intermediate frequency characteristics. In contrast, winner-take-all routing forces each subband to rely on a single expert, reducing routing flexibility and limiting the representation of mixed spectral patterns.

\noindent\textbf{The spectral descriptor and prompt play complementary roles (A12--A14).}
Prompt-only routing (A12, $\Delta{=}{-}0.26$\,dB) both degrades
performance and reorganizes routing around task identity rather than
frequency subspace.
Descriptor-only routing (A14, $-0.15$\,dB) retains more of the full
model's performance than prompt-only routing in this controlled setting,
showing that intrinsic spectral characteristics provide informative routing
cues. The remaining gap to the jointly conditioned gate indicates that
prompt-derived task context is complementary, particularly on deraining,
where degradation spectra can overlap.
A label-conditioned router that receives the ground-truth degradation
label (A13) is slightly stronger in distribution ($+0.06$\,dB on average
on BSD68; Table~\ref{tab:ood_routing}), confirming that frequency-based
routing is a competitive input-derived proxy for a signal WaFIR never
accesses at inference.

\noindent\textbf{Haar DWT and cross-subband interaction both contribute (A7, A9).}
Removing the grouped $1{\times}1$ cross-subband convolution
(A7, $-0.16$\,dB) confirms that inter-frequency interaction before IDWT
is necessary for coherent multi-scale reconstruction.
Replacing Haar DWT with average pooling (A9, $-0.10$\,dB) shows that
structured, lossless frequency decomposition is needed for the spectral
descriptor to access subband-specific statistics.

\noindent\textbf{Phase modulation is optional; magnitude modulation alone is
already sufficient (A4).}
Disabling phase modulation ($\varphi'{=}\varphi$) yields a marginal
improvement ($+0.04$\,dB average on the FULL metric), confirming that the
zero-initialized phase projector remains near-identity throughout training
and that the reported gains stem primarily from subband-resolved magnitude
modulation. All WaFIR numbers reported in the main paper use the full
configuration with phase modulation (A3) for consistency; the magnitude-only
variant (A4) is an equally strong and slightly lighter simplification.

\noindent\textbf{Load-balance regularization is essential (A8, A10--A11).}
Removing the load-balance loss (A8, $\lambda{=}0$) causes expert
collapse: expert utilization approaches $1.0$ and PSNR drops by
$-2.66$\,dB relative to the adopted $\lambda{=}0.01$ (A3) at a matched
epoch~105 on the fast-subset run (Table~\ref{tab:lb_ablation}).
The sweep over $\lambda$ establishes $\lambda{\in}[0.001, 0.01]$ as the
effective operating range; the full convergence analysis and formulation
rationale (comparison with Switch Transformer and entropy losses) are
given in Section~\ref{supp:lb}.

\noindent\textbf{Learned routing generalizes better than
label-conditioned routing under distribution shift (A3 vs.\ A13).}
The in-distribution comparison above assumes the evaluation condition
matches one of the five degradation labels seen at training time---an
assumption A13 depends on structurally, since its router selects experts
conditioned on a one-hot ground-truth label (A13 is thus not a true oracle
outside its label set). Table~\ref{tab:ood_routing}
stress-tests this assumption under three regimes: \emph{in-distribution}
(standard BSD68 noise levels), \emph{unseen intensity} (interpolated
$\sigma{=}35$ between trained $\sigma{\in}\{25,50\}$, and extrapolated
$\sigma{=}70$ beyond the trained range), and \emph{zero-shot compound
degradation} (two degradation types applied to the same image - e.g.\
rain$+$noise or haze$+$noise - never presented as a single label to
either model during training). A13's edge is confined to in-distribution
conditions ($+0.06$\,dB on average on BSD68); it \emph{inverts} once the
input falls outside the five trained labels: A3 outperforms A13 by
$+0.45$\,dB PSNR
($+0.046$ SSIM) at extrapolated $\sigma{=}70$, and by $+0.67$\,dB on
rain$+$noise. On haze$+$noise the two are statistically indistinguishable
($15.755$ vs.\ $15.739$\,dB) notably, \emph{both} models degrade
sharply on this condition (${\approx}12$--$18$\,dB below single-degradation
haze or noise performance), indicating compound haze$+$noise is a shared
failure mode rather than an advantage for either routing strategy.
The pattern that does hold is directional: whenever the ground-truth
label is unavailable or ill-defined for the input, A13 has no fallback
beyond its
five discrete classes, while A3's continuous frequency-domain descriptor
remains a well-defined signal regardless of the degradation's identity or
composition, making label-free frequency routing the more robust choice
when the deployed degradation distribution cannot be fully anticipated
at training time.

\begin{table}[t]
\centering
\caption{%
Out-of-distribution generalization, A3 (learned routing) vs.\ A13
(label-conditioned routing with the ground-truth degradation label). \emph{Unseen intensity} and
\emph{compound degradation} conditions are not seen as a single label
by either model during training. $\Delta = \text{A3} - \text{A13}$;
\textcolor{green!60!black}{green} favours A3,
\textcolor{red!70!black}{red} favours A13.
}
\label{tab:ood_routing}
\setlength{\tabcolsep}{3.5pt}
\resizebox{\linewidth}{!}{%
\begin{tabular}{l l c c r}
\toprule
Regime & Condition & A3 & A13 & $\Delta$ (PSNR) \\
\midrule
\multirow{3}{*}{In-distribution}
 & BSD68 $\sigma{=}15$ & 33.562/0.928 & 33.620/0.930 & \textcolor{red!70!black}{$-$0.058} \\
 & BSD68 $\sigma{=}25$ & 30.956/0.882 & 31.020/0.885 & \textcolor{red!70!black}{$-$0.064} \\
 & BSD68 $\sigma{=}50$ & 27.729/0.784 & 27.790/0.790 & \textcolor{red!70!black}{$-$0.061} \\
\midrule
\multirow{2}{*}{Unseen intensity}
 & $\sigma{=}35$ (interp.) & 29.308/0.841 & 29.265/0.840 & \textcolor{green!60!black}{$+$0.043} \\
 & $\sigma{=}70$ (extrap.) & \textbf{24.851}/\textbf{0.657} & 24.405/0.611 & \textcolor{green!60!black}{$+$0.446} \\
\midrule
\multirow{2}{*}{\makecell[l]{Compound\\degradation}}
 & Rain$+$noise$_{25}$ & \textbf{25.143}/\textbf{0.769} & 24.472/0.737 & \textcolor{green!60!black}{$+$0.671} \\
 & Haze$+$noise$_{25}$ & 15.755/0.706 & 15.739/0.707 & $+$0.016 (n.s.) \\
\bottomrule
\end{tabular}}
\end{table}

\subsection{Load-Balance Ablation}

Table~\ref{tab:lb_ablation} isolates the effect of $\mathcal{L}_\text{lb}$.
To avoid confounding regularization strength with training length, all four
settings (A8, A11, A10, A3; $\lambda\in\{0,10^{-4},10^{-3},10^{-2}\}$) are
evaluated at a matched epoch~105 on the fast-subset run, so any difference
reflects the regularization strength alone rather than training length.

At matched epoch~105, removing the loss (A8, $\lambda{=}0$) drops PSNR to
$25.81$\,dB versus $28.47$\,dB at the adopted $\lambda{=}0.01$ (A3) - a
\textbf{$-2.66$\,dB} gap attributable to regularization alone - with expert
utilization approaching $1.0$, confirming collapse onto a single pair. The
loss is thus not a minor regularizer but a precondition for the architecture
to function. It is also insensitive within a wide range: at epoch~105, the
working values $\lambda{=}0.0001$ (A11) and $\lambda{=}0.001$ (A10) lie within
$0.11$--$0.21$\,dB of the adopted $\lambda{=}0.01$, so any
$\lambda\in[10^{-4},10^{-2}]$ prevents collapse while $\lambda{=}0$ does not. We adopt $\lambda{=}0.01$ for the full model, which attains the best PSNR of
the four settings at epoch~105 ($28.471$\,dB, A3); this fast-subset,
epoch-105 value is not directly comparable to the 8-condition FULL score
($30.35$\,dB) reported for A3 in Table~\ref{stab:ablation}.

\begin{table}[h]
\centering
\caption{%
  Load-balance ablation. All four settings (A8/A11/A10/A3, i.e.\
  $\lambda\in\{0,10^{-4},10^{-3},10^{-2}\}$) are evaluated at a matched
  epoch~105 on the fast subset. $\Delta$PSNR is taken against A3, the adopted
  $\lambda{=}0.01$ configuration. Expert utilization $=\max_e\bar{w}_{s,e}$ averaged over
  test images and subbands; values near $1.0$ indicate collapse.
}
\label{tab:lb_ablation}
\resizebox{\linewidth}{!}{%
\begin{tabular}{llcccc}
\toprule
ID & $\lambda$ & PSNR & SSIM & $\Delta$\,PSNR & Expert util. \\
\midrule
\multicolumn{6}{l}{\emph{Matched epoch~105 (fast subset)}}\\
A8  & 0 (no lb)   & 25.812 & 0.8078 & \textcolor{red!70!black}{$-$2.659} & ${\approx}1.0$ \\
A11 & 0.0001      & 28.358 & 0.8695 & \textcolor{red!70!black}{$-$0.113} & -- \\
A10 & 0.001       & 28.259 & 0.8620 & \textcolor{red!70!black}{$-$0.212} & -- \\
A3  & 0.010       & \textbf{28.471} & \textbf{0.8720} & --- (ref.) & -- \\
\bottomrule
\end{tabular}}
\end{table}

\subsection{Seed Robustness}
\label{supp:seed}

To confirm that the component margins in Table~\ref{stab:ablation} exceed
random-initialization noise, we retrain the full model (A3) and the no-SPMoE
baseline (A0), each with two random seeds, and evaluate the best-validation
checkpoints (Table~\ref{tab:seed}). Across seeds, A3 attains FULL
$=30.29\pm0.02$\,dB and A0 $=29.83\pm0.04$\,dB; the $+0.46$\,dB SPMoE
improvement is consistent on both seeds ($+0.45$ and $+0.47$\,dB) and far
exceeds the per-model seed spread, confirming that the gain is architectural
rather than an initialization artifact. Because A3's seed deviation on FULL is
only $0.02$\,dB, the component deltas of $\geq0.14$\,dB in
Table~\ref{stab:ablation} all lie well outside seed noise. Frequency-structured
tasks are highly stable for A3 (BSD68 and Rain100L both $\pm0.02$\,dB),
whereas dehazing (SOTS) shows the largest seed spread ($\pm0.67$\,dB), matching the training-schedule instability discussed in Section~\ref{supp:limitations}.

\begin{table*}[h]
\centering
\footnotesize
\setlength{\tabcolsep}{4pt}
\renewcommand{\arraystretch}{0.95}
\caption{%
  Seed stability of the full model (A3) and the no-SPMoE baseline (A0),
  PSNR (dB). Two random seeds, best-validation checkpoints. BSD68 and Urban100
  are averaged over $\sigma{\in}\{15,25,50\}$; FULL is the mean over the eight
  conditions of Table~\ref{stab:ablation}. The $+0.46$\,dB A3$-$A0 gap on FULL
  is consistent across both seeds and exceeds the per-model seed spread.
}
\label{tab:seed}
\begin{tabular}{llccccc}
\toprule
Model & Seed & BSD68 & Urban100 & Rain100L & SOTS & FULL \\
\midrule
\multirow{3}{*}{A3}
 & 1 & 30.78 & 29.72 & 33.01 & 27.98 & 30.31 \\
 & 2 & 30.75 & 29.97 & 33.05 & 27.03 & 30.28 \\
 & mean$\pm$std
   & 30.76{\scriptsize$\,\pm.02$}
   & 29.84{\scriptsize$\,\pm.17$}
   & 33.03{\scriptsize$\,\pm.02$}
   & 27.50{\scriptsize$\,\pm.67$}
   & 30.29{\scriptsize$\,\pm.02$} \\
\midrule
\multirow{3}{*}{A0}
 & 1 & 30.63 & 29.46 & 31.66 & 26.92 & 29.86 \\
 & 2 & 30.62 & 29.41 & 31.32 & 27.03 & 29.81 \\
 & mean$\pm$std
   & 30.62{\scriptsize$\,\pm.01$}
   & 29.43{\scriptsize$\,\pm.04$}
   & 31.49{\scriptsize$\,\pm.24$}
   & 26.98{\scriptsize$\,\pm.08$}
   & 29.83{\scriptsize$\,\pm.04$} \\
\bottomrule
\end{tabular}
\end{table*}

\section{Additional Experimental Results}
\label{supp:additional_results}

This section provides the complete qualitative and quantitative evaluations summarized in the Experiments section of the main paper. Unless stated otherwise, all models are
evaluated without task-specific fine-tuning.

\subsection{Additional Qualitative Comparisons}
\label{supp:qual}

We provide qualitative comparisons grouped by task and by analysis type, so
that each figure can be read on its own; unless stated otherwise all models are
evaluated without task-specific fine-tuning, and panels with a reference image
are annotated with PSNR$\uparrow$/SSIM$\uparrow$/LPIPS$\downarrow$.

\begin{figure*}[t]
  \centering
  \includegraphics[width=\linewidth]{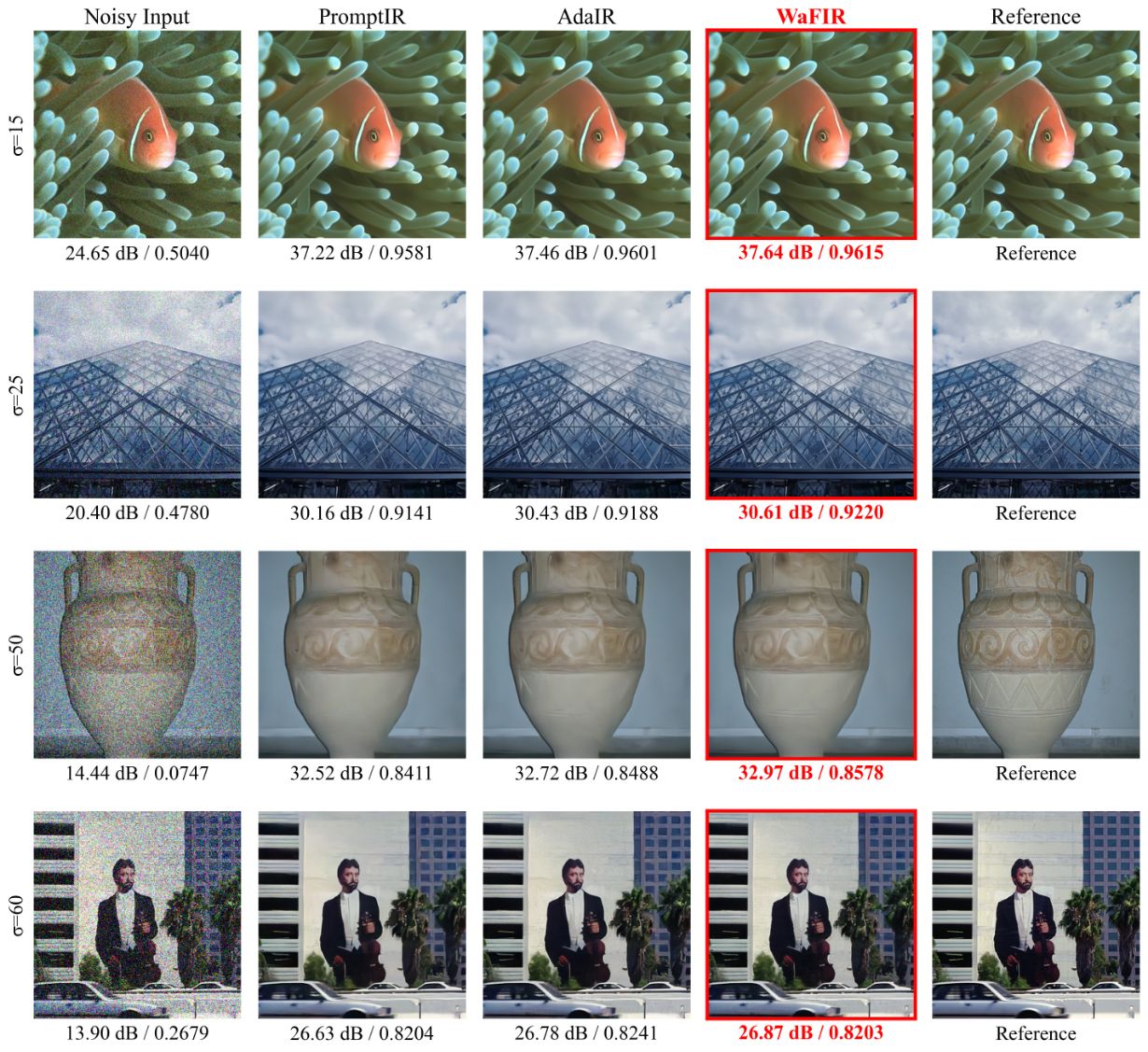}
  \caption{%
    \textbf{Qualitative denoising on BSD68.}
    Rows correspond to noise levels $\sigma\in\{15,25,50,60\}$.
    Columns show the noisy input, PromptIR, AdaIR, \textbf{WaFIR (ours)},
    and the reference image. WaFIR suppresses noise while preserving finer
    structures, with the differences becoming most visible at
    $\sigma\in\{50,60\}$. Zoomed crops highlight regions in which competing
    methods over-smooth texture or weaken edge contrast.
  }
  \label{fig:qual_denoise_bsd68}
\end{figure*}

\begin{figure*}[t]
  \centering
  \includegraphics[width=\linewidth]{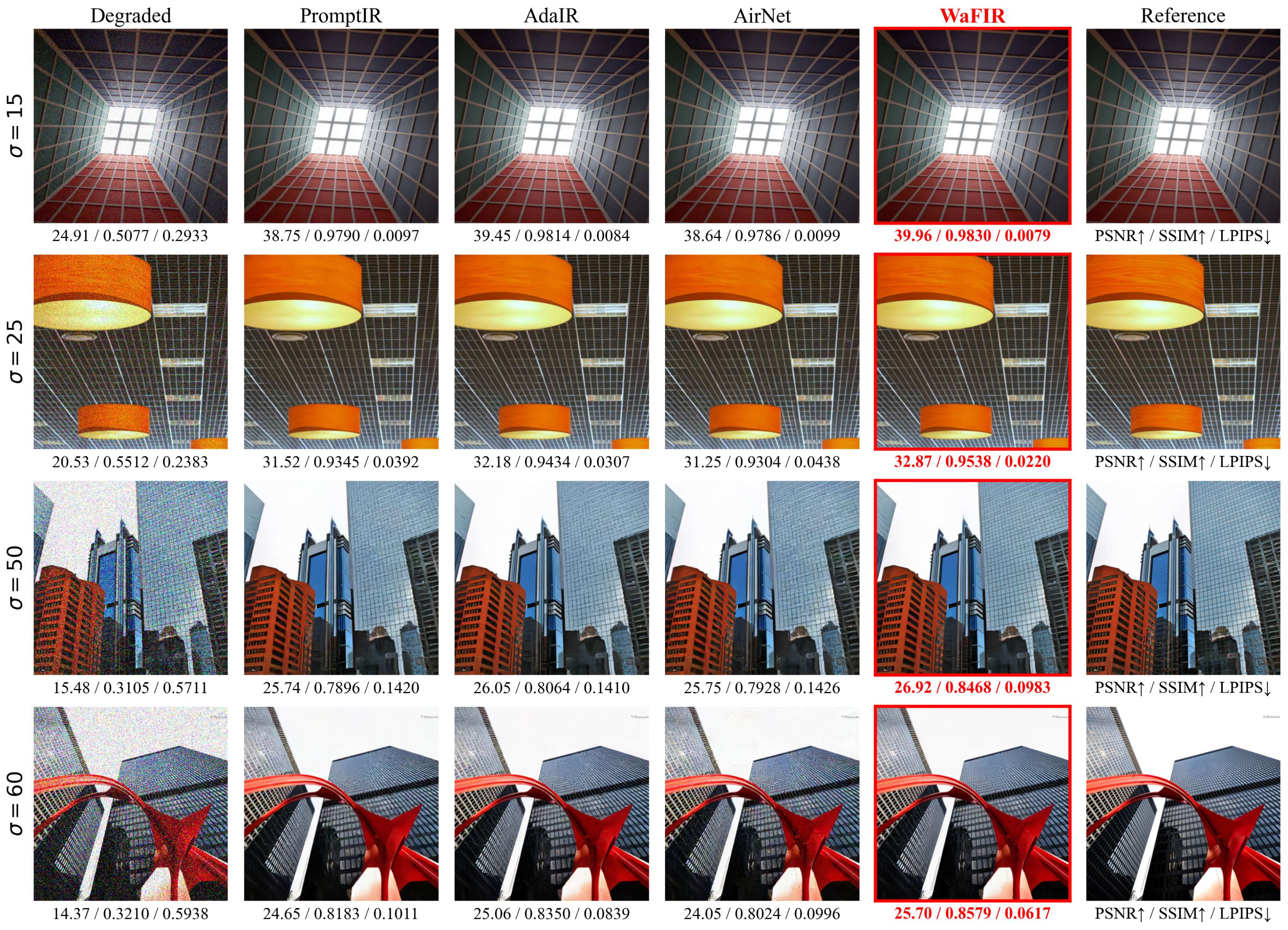}
  \caption{%
    \textbf{Cross-dataset denoising on Urban100.}
    The all-in-one models are evaluated directly on Urban100, including the
    out-of-range noise level $\sigma{=}60$, without fine-tuning.
    The comparison emphasizes preservation of repeated edges, line patterns,
    and high-frequency architectural details under dataset and noise-intensity
    shift.
  }
  \label{fig:qual_denoise_urban100}
\end{figure*}

\textbf{Denoising.} Figure~\ref{fig:qual_denoise_bsd68} compares methods on
BSD68 across noise levels $\sigma\in\{15,25,50,60\}$, and
Figure~\ref{fig:qual_denoise_urban100} on cross-dataset Urban100 including the
out-of-range $\sigma{=}60$; in both, WaFIR removes noise while preserving fine
texture and edge contrast, with the advantage most visible at high $\sigma$.

\begin{figure*}[t]
  \centering
  \includegraphics[width=\linewidth]{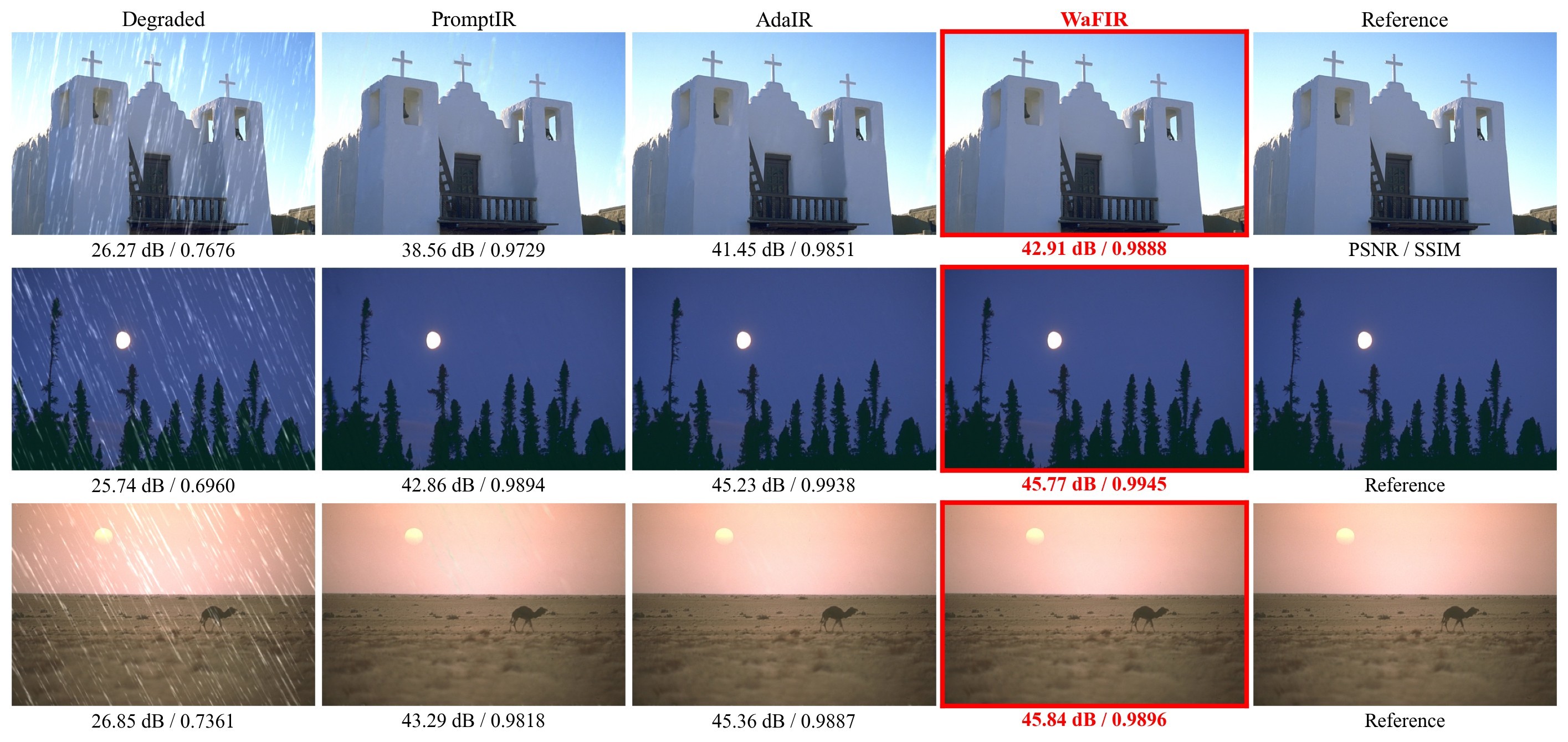}
  \caption{%
    \textbf{Qualitative deraining on Rain100L.}
    Columns show the rainy input, PromptIR, AdaIR, \textbf{WaFIR (ours)},
    and the reference image. WaFIR removes residual streaks while retaining
    fine scene structure and avoiding excessive smoothing.
  }
  \label{fig:qual_derain_rain100l}
\end{figure*}

\begin{figure*}[t]
  \centering
  \includegraphics[width=\linewidth]{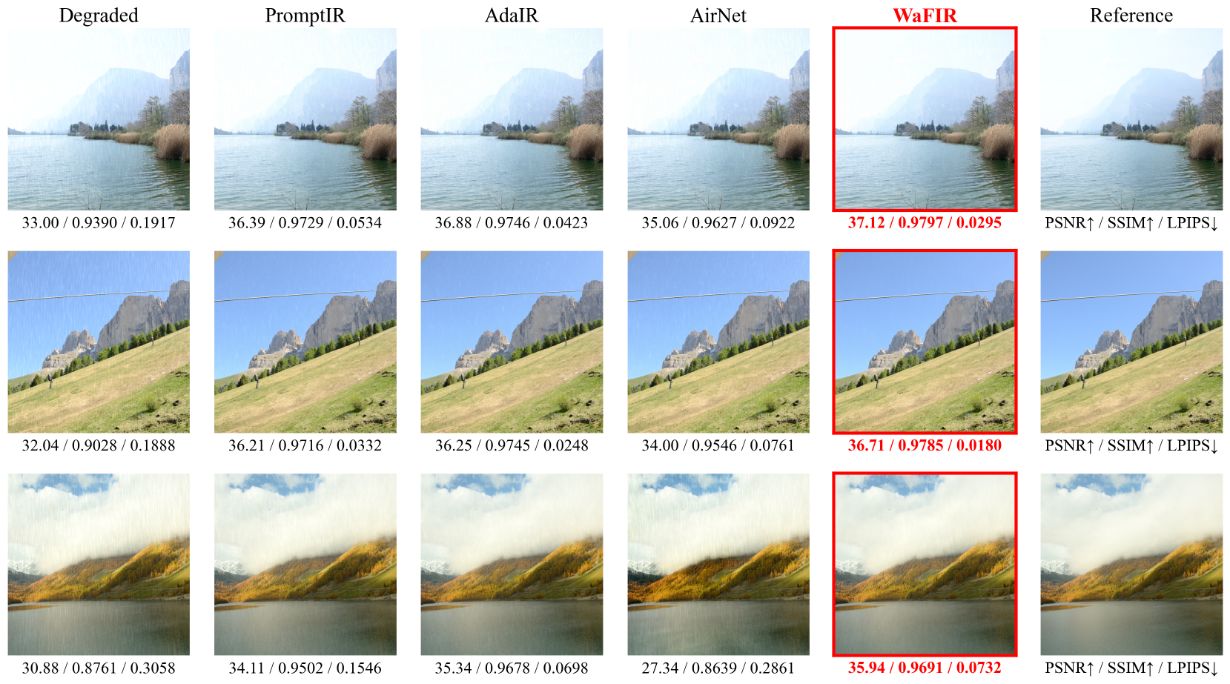}
  \caption{%
    \textbf{Qualitative deraining on the unseen CDD-11 dataset.}
    Models are trained on the source datasets and evaluated on CDD-11 without
    fine-tuning to test cross-dataset generalization. Columns show the rainy
    input, PromptIR, AdaIR, AirNet, \textbf{WaFIR (ours)}, and the reference;
    displayed metrics are PSNR$\uparrow$/SSIM$\uparrow$/LPIPS$\downarrow$.
    WaFIR removes rain streaks more thoroughly and attains the best PSNR/SSIM
    across the illustrated examples.
  }
  \label{fig:qual_derain_cdd}
\end{figure*}

\textbf{Deraining.} Figure~\ref{fig:qual_derain_rain100l} shows in-distribution
Rain100L, and Figure~\ref{fig:qual_derain_cdd} the unseen CDD-11 rain evaluated
without fine-tuning, where WaFIR removes residual streaks more completely while
keeping the underlying scene structure.

\begin{figure*}[t]
  \centering
  \includegraphics[width=\linewidth]{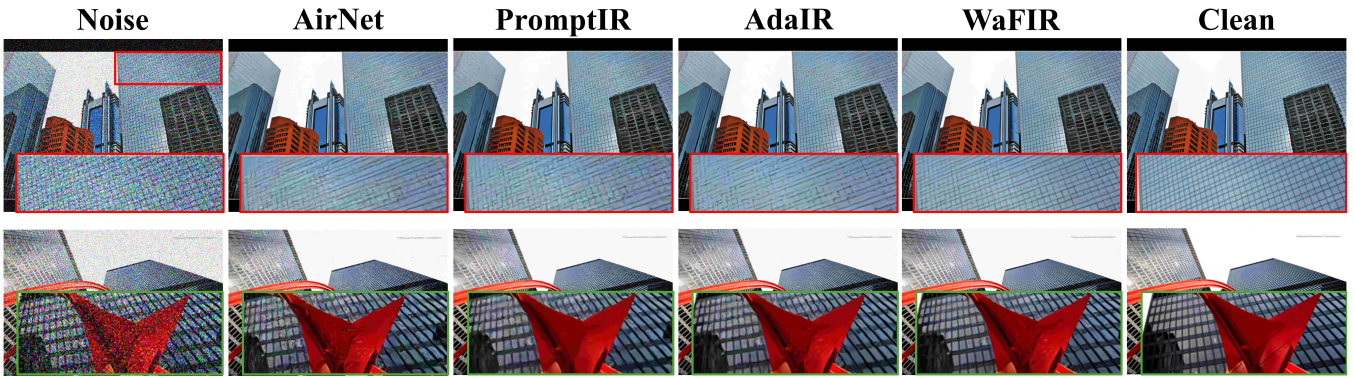}
  \caption{%
    \textbf{Zoomed line-structure recovery on Urban100.}
    Close-up crops of repeated line and lattice patterns under denoising.
    Because WaFIR modulates the \emph{global} Fourier magnitude within each
    subband, its correction is spatially coherent across the whole line
    structure; it therefore restores long, thin edges without the localized
    block or ringing artifacts that purely spatial (local-window) models tend
    to introduce along such structures. This behavior---coherent global
    frequency correction rather than piecewise local filtering---is a
    distinctive strength of WaFIR on periodic, high-frequency content.
  }
  \label{fig:zoom_line}
\end{figure*}

\begin{figure*}[t]
  \centering
  \includegraphics[width=\linewidth]{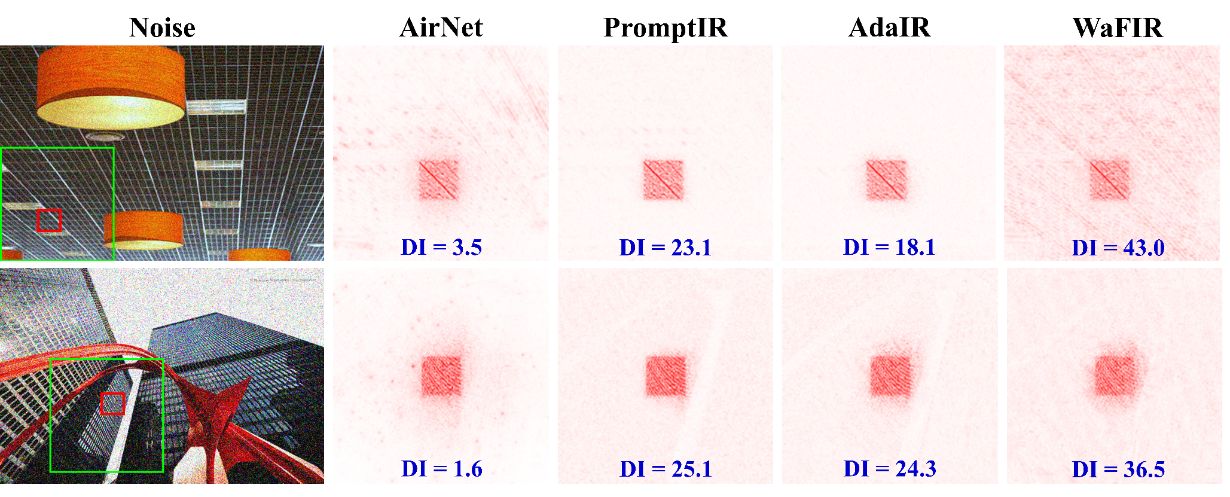}
  \caption{%
    \textbf{Local Attribution Maps (LAM) and Diffusion Index (DI) on
    Urban100.} For a fixed target region (red box), LAM~\citep{gu2021interpreting}
    highlights the input pixels that contribute to reconstructing that region;
    the Diffusion Index (DI) summarizes the spatial extent of this attribution
    (higher DI $=$ information aggregated from a wider area). The DI value is
    annotated below each panel. WaFIR attains the highest DI on both examples
    (row~1: $43.0$ vs.\ AdaIR $18.1$, PromptIR $23.1$, AirNet $3.5$; row~2:
    $36.5$ vs.\ AdaIR $24.3$, PromptIR $25.1$, AirNet $1.6$), exhibiting a
    markedly broader attribution footprint, whereas the spatial baselines
    concentrate attribution in a narrower neighborhood. This is the direct
    interpretability signature of WaFIR's design: subband-wise \emph{global}
    Fourier modulation lets every output pixel draw on long-range,
    structure-consistent frequency evidence, which explains both the
    artifact-free line recovery in Figure~\ref{fig:zoom_line} and the sharper
    edges reported quantitatively in Table~\ref{stab:ood}.%
  }
  \label{fig:lam_di}
\end{figure*}

\begin{figure*}[t]
  \centering
  \includegraphics[width=\linewidth]{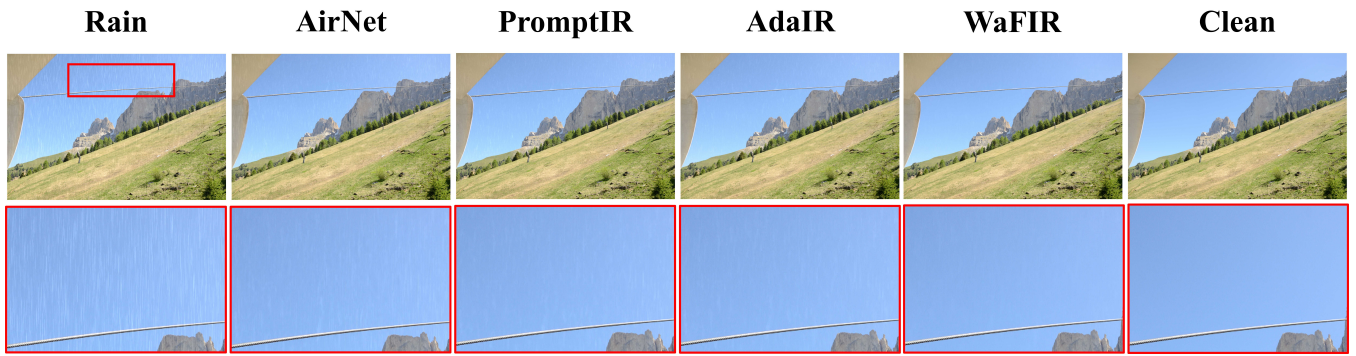}
  \caption{%
    \textbf{Zoomed deraining on the unseen CDD-11 dataset (out-of-distribution).}
    Close-up crops for a rain pattern outside the training distribution.
    Evaluated without fine-tuning, WaFIR removes residual streaks more
    completely while preserving underlying texture, illustrating that
    subband-resolved frequency routing generalizes to rain statistics never
    seen during training.
  }
  \label{fig:zoom_rain_ood}
\end{figure*}

\textbf{Structure preservation and interpretability.}
Figures~\ref{fig:zoom_line}, \ref{fig:lam_di}, and \ref{fig:zoom_rain_ood} zoom
into thin, repeated structures---line and lattice recovery on Urban100, Local
Attribution Maps with the Diffusion Index (DI), and out-of-distribution
rain-streak removal, respectively---and motivate the quantitative structure
analysis in Section~\ref{supp:line_rain}. Taken together, the figures show that
WaFIR's subband-resolved frequency modulation yields spatially coherent
corrections along extended structures rather than piecewise local filtering.

\subsection{Effectiveness on Line Artifacts and Rain Structures}
\label{supp:line_rain}

A recurring failure mode of purely spatial restorers is that they process each
pixel from a local window, so on long, thin, repeated structures they either
leave residual artifacts or over-smooth the structure. WaFIR modulates the
\emph{global} Fourier magnitude \emph{within each subband}, so its correction
is spatially coherent along an entire structure rather than filtered
piecewise. We examine the two structure-dominated cases directly.

\paragraph{Line structures (Urban100).}
On the repeated line and lattice patterns of Urban100 (Figure~\ref{fig:zoom_line}),
the spatial baselines introduce localized ringing or block artifacts along thin
edges, whereas WaFIR restores the full line coherently without such artifacts.
Because a periodic line structure maps to a compact, oriented signature in the
frequency domain, subband-wise magnitude modulation corrects the whole
structure at once instead of reconstructing it segment by segment precisely
where local-window models break the line into artifacts.

\paragraph{Rain streaks (out-of-distribution CDD-11).}
Rain streaks concentrate directional energy in the LH/HL subbands. Evaluated
without fine-tuning on the unseen CDD-11 rain (Figure~\ref{fig:zoom_rain_ood}),
WaFIR removes residual streaks more completely while preserving the underlying
texture, showing that subband-resolved routing transfers to rain statistics not
seen during training.

\paragraph{Attribution analysis (LAM / Diffusion Index).}
We quantify \emph{why} using Local Attribution Maps
(LAM)~\citep{gu2021interpreting} and their Diffusion Index (DI), which measures
how widely the contributing input pixels are spread (Figure~\ref{fig:lam_di}).
WaFIR shows a substantially broader attribution footprint and the highest DI on
both examples ($43.0$ and $36.5$, versus $18.1$--$25.1$ for the prompt/frequency
baselines and $\leq3.5$ for AirNet): each output pixel draws on long-range,
structure-consistent frequency evidence rather than a local window. This wide,
frequency-coherent attribution field is the mechanistic explanation for the
artifact-free line and rain recovery above, and directly reflects the design
thesis, subband-wise \emph{global} Fourier modulation trades the local
receptive field of purely spatial models for a wide, structure-aware one.

\subsection{Out-of-Distribution Evaluation}
\label{supp:ood_results}

\begin{table}[t]
\centering
\setlength{\tabcolsep}{4pt}
\footnotesize
\caption{\textbf{Out-of-distribution generalization.} The all-in-one model
(trained on Noise–Haze–Rain) is tested \emph{directly} (no fine-tuning) on: an unseen
rain dataset (CDD-11), unseen natural-image denoising (Urban100), and an
\emph{industrial domain}—Scanning Acoustic Microscopy (SAM) for electronic
semiconductor inspection (synthetic AWGN). Each setting includes an out-of-range
noise level $\sigma{=}60$ beyond the training regime. Metrics:
PSNR$\uparrow$/SSIM$\uparrow$/LPIPS$\downarrow$; best per block in \textbf{bold}.}
\label{stab:ood}
\begin{tabular}{@{}llccc@{}}
\toprule
Setting & Method & PSNR$\uparrow$ & SSIM$\uparrow$ & LPIPS$\downarrow$ \\
\midrule
\multirow{4}{*}{\shortstack[l]{CDD-11\\(rain)}}
 & PromptIR & 24.44 & 0.8638 & 0.1734 \\
 & AdaIR    & 24.73 & 0.8625 & 0.1691 \\
 & AirNet   & 23.07 & 0.8263 & 0.2198 \\
 & \textbf{WaFIR} & \textbf{24.94} & \textbf{0.8746} & \textbf{0.1577} \\
\midrule
\multirow{4}{*}{\shortstack[l]{Urban100\\avg\\($\sigma\!\in\!15\text{-}50$)}}
 & PromptIR & 30.77 & 0.9121 & 0.0612 \\
 & AdaIR    & 31.23 & 0.9193 & 0.0569 \\
 & AirNet   & 30.77 & 0.9135 & 0.0592 \\
 & \textbf{WaFIR} & \textbf{31.51} & \textbf{0.9249} & \textbf{0.0513} \\
\midrule
\multirow{4}{*}{\shortstack[l]{Urban100\\$\sigma{=}60$\\(OOD)}}
 & PromptIR & 26.36 & 0.8219 & 0.1137 \\
 & AdaIR    & 26.65 & \textbf{0.8229} & \textbf{0.1046} \\
 & AirNet   & 24.95 & 0.7411 & 0.1826 \\
 & \textbf{WaFIR} & \textbf{26.78} & 0.8148 & 0.1061 \\
\midrule
\multirow{4}{*}{\shortstack[l]{SAM\\avg\\($\sigma\!\in\!25\text{-}50$)}}
 & PromptIR & 31.42 & 0.8108 & 0.2859 \\
 & AdaIR    & 31.52 & 0.8128 & 0.2936 \\
 & AirNet   & 31.32 & 0.8049 & 0.3243 \\
 & \textbf{WaFIR} & \textbf{31.63} & \textbf{0.8152} & \textbf{0.2718} \\
\midrule
\multirow{4}{*}{\shortstack[l]{SAM\\$\sigma{=}60$\\(OOD)}}
 & PromptIR & 29.15 & \textbf{0.7320} & 0.3100 \\
 & AdaIR    & \textbf{29.20} & 0.7317 & 0.2492 \\
 & AirNet   & 27.09 & 0.6509 & 0.3798 \\
 & \textbf{WaFIR} & 29.16 & 0.7264 & \textbf{0.1904} \\
\bottomrule
\end{tabular}
\end{table}

\begin{figure*}[t]
  \centering
  \includegraphics[width=\linewidth]{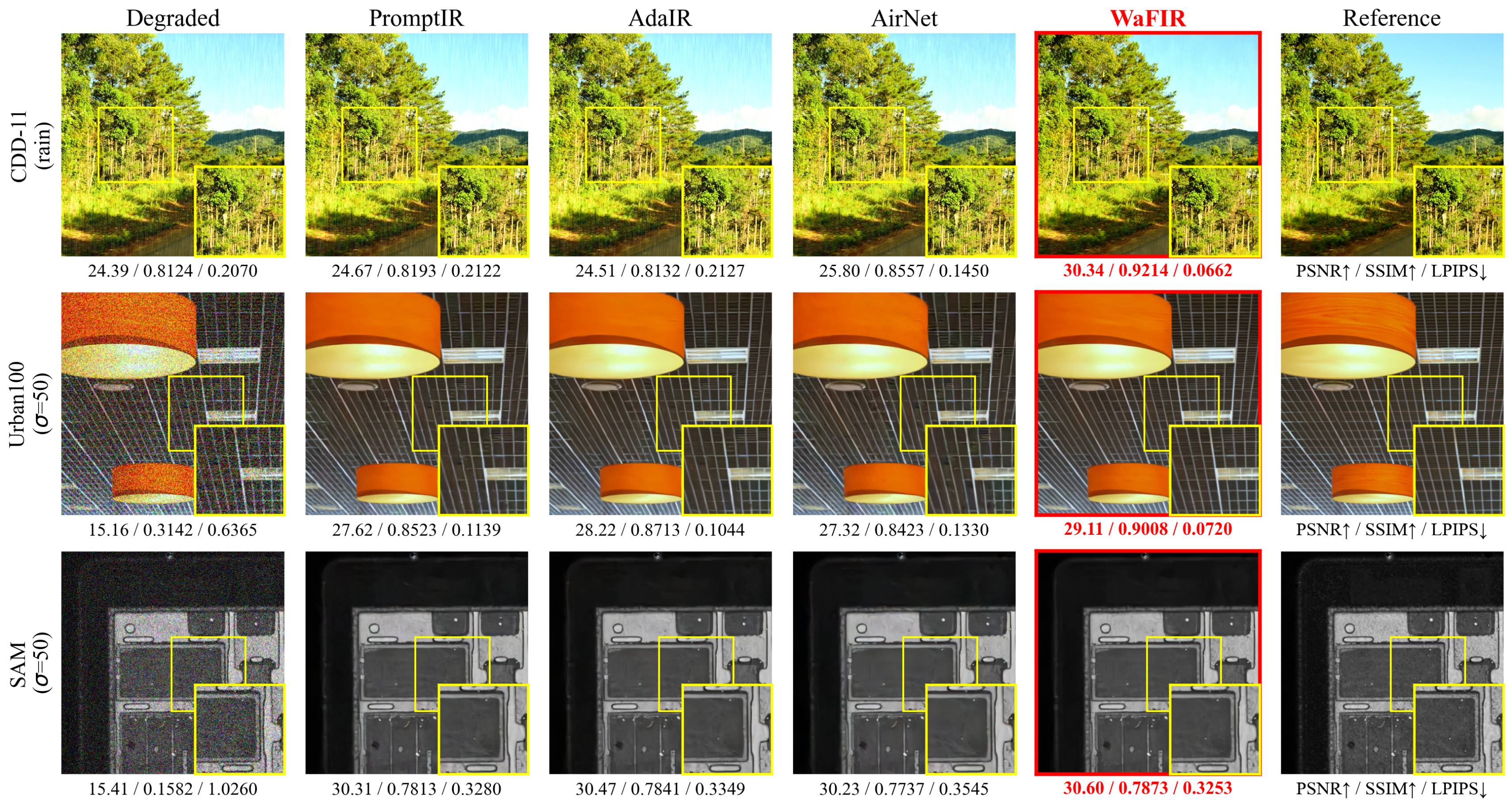}
 \caption{\textbf{Qualitative cross-domain generalization.} One example per domain:
unseen rain (CDD-11), natural-image denoising (Urban100, $\sigma{=}50$), and an
industrial domain—Scanning Acoustic Microscopy of an electronic semiconductor
package (SAM, $\sigma{=}50$). Columns: degraded input, PromptIR, AdaIR, AirNet,
WaFIR (ours), reference; PSNR$\uparrow$/SSIM$\uparrow$/LPIPS$\downarrow$ below each
panel. Trained only on noise--haze--rain and applied without fine-tuning,
WaFIR attains the best metrics on the examples shown; per-metric outcomes
at $\sigma{=}60$ vary across methods (Table~\ref{stab:ood}).}
  \label{fig:ood_domains}
\end{figure*}

\paragraph{For out-of-distribution evaluation:} Figures~\ref{fig:qual_derain_cdd} and \ref{fig:zoom_rain_ood} together with Table~\ref{stab:ood} present representative results comparing WaFIR against state-of-the-art all-in-one methods.

\section{Computational Complexity}
\label{supp:complexity}

Table~\ref{tab:params} summarizes the parameter overhead of the three
SPMoE levels. The deepest level (spmoe3, $C{=}384$) dominates due to
the wider channel dimension. The total overhead is $+6.56$M parameters
($+18.4\%$) over the PromptIR baseline.

\begin{table}[h]
\centering
\caption{Parameter breakdown by SPMoE level.}
\label{tab:params}
\begin{tabular}{lcc}
\toprule
Component & Channels $C_\ell$ & Params \\
\midrule
PromptIR baseline & -- & 35.59M \\
SPMoE-L1 (spmoe1) & 96  & $+$0.32M \\
SPMoE-L2 (spmoe2) & 192 & $+$1.22M \\
SPMoE-L3 (spmoe3) & 384 & $+$5.02M \\
\midrule
\textbf{WaFIR (full)} & -- & \textbf{42.15M} \\
\bottomrule
\end{tabular}
\end{table}

We further compare WaFIR against representative all-in-one restoration
baselines in terms of parameter count, computational cost, and wall-clock
inference latency. Table~\ref{tab:complexity} reports results measured on
the CDD dataset test set (200 rain images), on a single RTX 3060 GPU.
AirNet has the fewest parameters (8.93M), but the highest FLOPs (602.55G)
and the slowest inference, since its contrastive encoder processes the
input twice and its deformable convolutions are computationally expensive.
AdaIR is the most parameter- and compute-efficient baseline (28.79M
params, 294.36G FLOPs). WaFIR adds approximately 6M parameters over
PromptIR for the spectral mixture-of-experts decoder, with only a modest
increase in FLOPs and inference latency, while achieving the best
restoration quality. 

\paragraph{Time and space complexity.}
With $N{=}HW$ spatial positions and $C$ channels, each SPMoE block is dominated
by the per-subband FFT/iFFT at $\mathcal{O}(NC\log N)$~\cite{cooley1965} and the
$1{\times}1$ channel mixing at $\mathcal{O}(NC^{2})$, giving
$\mathcal{O}(NC\log N + NC^{2})$ time, i.e.\ near-linear in $N$ and free of the
$\mathcal{O}(N^{2}C)$ cost of spatial self-attention. Activation memory is
$\mathcal{O}(NC)$ (linear in $N$), and the added parameters are
$\mathcal{O}(C^{2})$ per level, independent of resolution ($6.56$M total,
Table~\ref{tab:params}).

\begin{table}[h]
\centering
\caption{Computational complexity and restoration quality comparison,
measured on the CDD dataset test set (200 rain images) on a single
RTX 3060 GPU. Inference time is reported as mean $\pm$ std (ms);
PSNR/SSIM are averaged over the same 200 images.}
\label{tab:complexity}
\resizebox{\columnwidth}{!}{%
\begin{tabular}{lcccc}
\toprule
Model & Params (M) & GFLOPs & Inference (ms) & PSNR / SSIM \\
\midrule
PromptIR & 35.59 & 316.28 & 180.87 $\pm$ 0.37 & 24.4413 / 0.8638 \\
AdaIR    & 28.79 & 294.36 & 190.44 $\pm$ 2.96 & 24.7290 / 0.8625 \\
AirNet   &  8.93 & 602.55 & 320.97 $\pm$ 2.69 & 23.0681 / 0.8263 \\
WaFIR    & 42.15 & 321.11 & 199.52 $\pm$ 1.63 & \textbf{24.9357} / \textbf{0.8746} \\
\bottomrule
\end{tabular}%
}
\end{table}

\section{Limitations and Future Work}
\label{supp:limitations}

Our model has several limitations that should be acknowledged.

\begin{itemize}
\item \textbf{Parameter overhead.} The $+$18.4\% parameter overhead (42.15M total) places WaFIR above
lightweight competitors (AnyIR~\cite{anyir}: 6M, MoCE-IR~\cite{moceir}: 25M). Two directions could reduce
this cost. The first is knowledge distillation from WaFIR to a compact student.
The second exploits the interpolation properties of the spectral (Fourier)
domain: since the per-frequency magnitude and phase modulations produced by the
experts vary smoothly across frequency, they could be predicted on a coarse
frequency grid and interpolated back to full resolution, so that the spectral
experts operate on a reduced set of frequency coefficients and cut both
parameters and compute.

\item \textbf{Dehazing.} WaFIR trails its PromptIR reference on dehazing
(29.70 vs.\ 30.58\,dB on SOTS), despite identical full-RESIDE training. We
attribute this to the degradation and the training schedule, not the SPMoE
module. Under the atmospheric scattering model $I{=}Jt{+}A(1{-}t)$~\cite{he2009single},
haze is a smooth, orientation-free, low-frequency degradation that lacks the
fixed directional signature rain and noise provide, so the subband router has
little structure to key on. Empirically, the shared schedule also under-serves
dehazing (repeat factor one, versus heavily up-sampled rain and noise), yet
under a matched budget SPMoE still \emph{improves} SOTS over PromptIR
(27.56 vs.\ 27.06\,dB), indicating an optimization-balance effect rather than an
architectural limit. Two directions could close the gap: task-balanced
scheduling (e.g., uncertainty-based weighting), and adaptively up-weighting the
prompt context when the spectral descriptor is uninformative, as for haze.

\item \textbf{Degradation scope and compound cases.} WaFIR currently covers noise, rain, and haze;
extending to deblurring and real-world images is a natural direction. Compound
degradations further remain a shared failure mode: on haze$+$noise, both
learned and label-conditioned routing collapse to ${\approx}15.7$\,dB
(Table~\ref{tab:ood_routing}), and at $\sigma{=}60$ WaFIR does not lead on
every metric (Table~\ref{stab:ood}); robust mixed-degradation handling remains
open.

\item \textbf{Spectral descriptor.} The amplitude-invariance control in Section~\ref{supp:amplitude} suggests
that ratio-based spectral descriptors (D1) may outperform the current
absolute-magnitude descriptor. Substituting D1 and measuring the impact on
PSNR is a concrete future experiment.
\end{itemize}

\end{document}